\documentclass{ecai} 
\usepackage{latexsym}
\usepackage{amssymb}
\usepackage{amsmath}
\usepackage{amsthm}
\usepackage{booktabs}
\usepackage{enumitem}
\usepackage{graphicx}
\usepackage{color}
\usepackage{graphicx}
\usepackage{booktabs}  % 用于制作更精美的表格
\usepackage[table,xcdraw]{xcolor} % 引入颜色支持
\usepackage{multirow} % 支持跨行合并
\usepackage{subfigure}
\usepackage{pifont}
\usepackage{algorithm}
\usepackage{algorithmic}
\usepackage{cleveref}
\usepackage{wasysym}

\newcommand{\BibTeX}{B\kern-.05em{\sc i\kern-.025em b}\kern-.08em\TeX}

\begin{document}

%%%%%%%%%%%%%%%%%%%%%%%%%%%%%%%%%%%%%%%%%%%%%%%%%%%%%%%%%%%%%%%%%%%%%%%%

\begin{frontmatter}

%%% Use this command to specify your submission number.
%%% In doubleblind mode, it will be printed on the first page.

% \paperid{6968} 

%%% Use this command to specify the title of your paper.

% \title{FedMate: Facilitating Complementary Knowledge Fusion in Federated Learning via Meticulously-calibrated Aggregation and Merit-discrimination Training}
\title{Choice Outweighs Effort: Facilitating Complementary Knowledge Fusion in Federated Learning via Re-calibration and Merit-discrimination}

%%% Use this combinations of commands to specify all authors of your 
%%% paper. Use \fnms{} and \snm{} to indicate everyone's first names 
%%% and surname. This will help the publisher with indexing the 
%%% proceedings. Please use a reasonable approximation in case your 
%%% name does not neatly split into "first names" and "surname".
%%% Specifying your ORCID digital identifier is optional. 
%%% Use the \thanks{} command to indicate one or more corresponding 
%%% authors and their email address(es). If so desired, you can specify
%%% author contributions using the \footnote{} command.

% \author[A]{\fnms{First}~\snm{Author}\orcid{....-....-....-....}\thanks{Corresponding Author. Email: somename@university.edu.}\footnote{Equal contribution.}}
% \author[B]{\fnms{Second}~\snm{Author}\orcid{....-....-....-....}\footnotemark}
% \author[B,C]{\fnms{Third}~\snm{Author}\orcid{....-....-....-....}} 

% \address[A]{Short Affiliation of First Author}
% \address[B]{Short Affiliation of Second Author and Third Author}
% \address[C]{Short Alternate Affiliation of Third Author}

\author[A,B]{\fnms{Ming}~\snm{Yang}}
\author[A,B]{\fnms{Dongrun}~\snm{Li}}
\author[A,B]{\fnms{Xin}~\snm{Wang}\thanks{Corresponding author. Email: xinwang@qlu.edu.cn.
% \\Code: \url{https://github.com/Dongrun-Li/FedMate.git}.
% \\Full version of this paper can be found in [A].
}
}
\author[A,B]{\fnms{Xiaoyang}~\snm{Yu}}
\author[A,B]{\fnms{Xiaoming}~\snm{Wu}}
\author[C]{\fnms{Shibo}~\snm{He}}

\address[A]{Key Lab. of Computing Power Network and Information Security, Ministry of Education,
Shandong Computer Science Center, Qilu University of Technology (Shandong Academy of Sciences), Jinan, China}

\address[B]{Shandong Provincial Key Lab. of Industrial Network and Information System Security,
Shandong Fundamental Research Center for Computer Science, Jinan, China}

\address[C]{College of Control Science and Engineering, Zhejiang University, Hangzhou, China}

%%% Use this environment to include an abstract of your paper.

\begin{abstract}
Cross-client data heterogeneity in federated learning induces biases that impede unbiased consensus condensation and 
% the fusion of complementary knowledge for generalization and personalization.
the complementary fusion of generalization- and personalization-oriented knowledge. 
While existing approaches mitigate heterogeneity through model decoupling and representation center loss, they often rely on static and restricted
%to enable fine-grained personalized FL, partially alleviating heterogeneity-related challenges.
%However, these approaches typically rely on static and restricted 
% myopic 
metrics to evaluate local knowledge and adopt global alignment too rigidly, leading to consensus distortion and diminished model adaptability. 
To address these limitations, we propose \textit{FedMate}\footnotemark[1], a method that implements bilateral optimization: On the server side, we construct a dynamic global prototype, with aggregation weights calibrated by 
holistic integration of sample size, current parameters, and future prediction; a category-wise classifier is then fine-tuned using this prototype to preserve global consistency. 
% across the full lifecycle. %by jointly considering historical training, current parameters, and future usage. holistic/comprehensive? 
% and maintain global consistency by fine-tuning the classifier with global prototypes. 
On the client side, we introduce complementary classification fusion to enable merit-based discrimination training and incorporate cost-aware feature transmission to balance model performance and communication efficiency. 
Experiments on five datasets of varying complexity demonstrate that FedMate outperforms state-of-the-art methods in harmonizing generalization and adaptation. Additionally, semantic segmentation experiments on autonomous driving datasets validate the method's real-world scalability. 
%We compared FedMate with 13 classical and SOTA approaches 
% Experiments across five datasets of differing complexities to evaluate its generalization and adaptation. FedMate exhibited notable performance. To evaluate its real-world applicability, we carried out semantic segmentation experiments using autonomous driving datasets. The impressive results further underscore the scalability of FedMate.
\end{abstract}

\end{frontmatter}

\textbf{Keyworks:} Personalized Federated Learning, Heterogeneous Data, Complementary Knowledge Fusion
\footnotetext[1]{Code: \url{https://github.com/Dongrun-Li/FedMate.git}.
% \\Full version of this paper can be found in [A].
}
\section{Introduction}\label{intro}
During artificial intelligence (AI) model training, participants often seek to enhance their models by leveraging others’ data while unwilling to share their own private and high-value information~\cite{liang2022advances}. Federated learning (FL) addresses this dilemma by facilitating the exchange of locally trained models instead of raw data~\cite{mcmahan2017communication}, thereby enabling secure knowledge sharing and collaborative training across distributed clients. 
A streamlined implementation framework allows FL to be seamlessly applied to lightweight models~\cite{qi2024small}. Recent work, such as Google’s DiLoCo framework~\cite{charles2025communication}, demonstrates the scalability of FL for large language models, reinforcing its effectiveness in distributed training. Similarly, frameworks integrating general-purpose and specialized models align naturally with FL principles~\cite{cheng2022fedgems,yu2023multimodal}, underscoring FL’s versatility.
%the DiLoCo framework in their research on large language model scaling laws, which aligns conceptually with FL, further validating its effectiveness in large-scale distributed training. Additionally, the large and small models collaboration framework, designed to integrate general-purpose and specialized models, naturally resonates with FL~\cite{cheng2022fedgems,yu2023multimodal}. These developments highlight the exceptional potential of FL.

However, in real-world applications, the decentralized nature of FL data exacerbates cross-client data heterogeneity, hindering the single global model's ability to adapt to diverse local tasks~\cite{ye2023heterogeneous}. 
% ,zhang2021survey
Consequently, achieving an effective balance between model generalization and personalization has emerged as a key research challenge. 
To address this, personalized federated learning (PFL) has been introduced, aiming to optimize this trade-off through fine-grained methodologies~\cite{collins2021exploiting,oh2022fedbabu,chen2022on,xu2023personalized}. 
% tan2022towards,
PFL typically involves two components: personalized aggregation on the server and personalized training on the client. The latter often proves more effective due to greater resource availability and operational flexibility, enabling closer alignment with the desired objectives. 
We argue that high-quality local training hinges critically on two factors: selecting personalized modules within the local model and designing both the method and intensity of generalization constraints~\cite{zhang2023gpfl,deng2020adaptive}.
\begin{figure}[!t]
    \centering
    \includegraphics[width=0.47\textwidth]{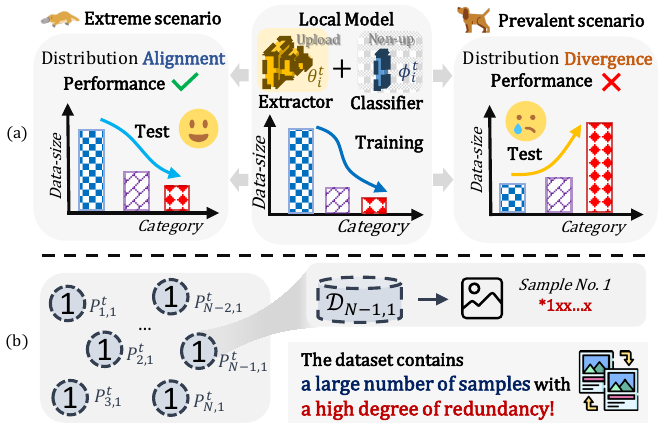}
    \caption{
    Illustration of the deficiencies in existing FL approaches regarding model decoupling and class center loss design.
    % Limitations of existing FL methods in model decoupling and class center loss design.
    }
    \label{fig:limit}
\end{figure}
Decoupling a model into a shared feature extractor and a task-specific classifier provides a promising approach to modular personalization in PFL~\cite{arivazhagan2019federated,collins2021exploiting}. 
This architecture leverages the transferable nature of feature representations: a globally aggregated feature extractor (e.g., via FedAvg~\cite{mcmahan2017communication}) can effectively capture generalizable features across clients, even under data distribution shifts. In contrast, classifiers---being more sensitive to local task distributions---typically require personalization to adapt to client-specific variations.
However, current methods adopt an extreme dichotomy in optimization: they enforce rigid generalization on the feature extractor while allowing unconstrained personalization of classifiers. This strict decoupling harms component compatibility, ultimately degrading overall model performance. As illustrated in Fig.~\ref{fig:limit}(a), fully personalized classifiers are prone to overfitting, developing biased decision boundaries that limit their operational scope. 
%However, existing approaches often treat the training objectives of the decoupled components in a fully disjointed manner—imposing strong generalization constraints on the feature extractor while applying full personalization to the classifier.
%This disconnection undermines the compatibility between the two components, leading to suboptimal overall model performance. Moreover, as illustrated in Fig.~\ref{fig:limit}(a), fully personalized classifiers tend to overfit to local distributions, resulting in overly biased decision boundaries and constraining the model’s effectiveness within a narrow operational range.
These limitations arise from: insufficient global knowledge integration during local training, and the lack of a % balanced 
granular trade-off mechanism within each decoupled component. 
%This issue primarily stems from the inadequate incorporation of global knowledge during local training, compounded by the need to maintain an internal trade-off within each decoupled component.
Thus, the core challenge is: \textit{How can decoupled components dynamically incorporate complementary generalization-personalization knowledge during training?} 
%Accordingly, the first key challenge lies in: \textit{how can decoupled components effectively integrate complementary generalization-personalization knowledge during model training?} 

Leveraging high-quality global knowledge at the feature level enhances generalization by promoting consistent and transferable feature representations across distributed clients. Among existing methods to achieve this, center loss has demonstrated strong potential for  improving feature discriminability by enforcing intra-class compactness and inter-class separability~\cite{wen2016discriminative, florea2020margin,liu2024vmamba}.
% Center loss has demonstrated significant potential for learning discriminative feature spaces by enforcing intra-class compactness and inter-class separability~\cite{wen2016discriminative,abramson2024accurate,liu2024vmamba}. 
In typical implementations, global representation anchors are employed to impose center loss constraints on local features prior to logits-related cross-entropy optimization. 
%Prior to conducting local training with logits-related cross-entropy loss, global representation anchors are utilized to construct a center loss constraint on the extracted features. 
Nevertheless, existing approaches often overlook a critical prerequisite: validating the reliability of these global anchors before initiating global-guided training.
%ensuring the validity of global knowledge before initiating global-guided training is essential, yet this critical step is frequently neglected in existing methodologies. 
Furthermore, during server-side model aggregation, current methods primarily rely on local sample size as a proxy for assessing client update quality. 
%In assessing the validity of locally uploaded information, current methods often consider only the local training sample size as the metric of effectiveness. 
As shown in Fig.~\ref{fig:limit}(b), this simplistic metric proves inadequate for ensuring robust global consensus. 
This reveals the second fundamental challenge: \textit{How can the server evaluate client updates fairly during aggregation to derive an unbiased global consensus?} 

Building upon these two challenges, our key contributions are: 
 % \vspace{-3pt}
\begin{itemize}
    \item We propose FedMate, a novel PFL framework that strengthens global guidance through aggregation re-calibration to ensure unbiased consensus, while mutually refining supervisory signals to maintain consistency. Simultaneously, FedMate incorporates discriminative local selection, preserving complementary knowledge between generalization and personalization objectives.
    %\item We propose FedMate, a novel PFL framework that implements bilateral optimization: dynamically calibrates aggregation weights by holistically integrating sample size, current parameters, and future prediction to establish an unbiased global consensus. During local training, FedMate performs discriminative selection to retain beneficial knowledge from global-local information.
    \item We implement bilateral optimization on both server and client sides. On the server side, multi-view prototype scrutiny (MPS) and fine-grained category-wise classifier integration (CCI) are employed to ensure robust global aggregation. On the client side, we introduce complementary classification fusion (CCF) to reconcile global coherence and local adaptability, and cost-aware feature transmission (CFT) to regulate communication overhead.
    % \item Concretely, the server-side employs {Multi-view Prototype Scrutiny} (MPS) and fine-grained {Category-wise Classifier Integration} (CCI), enabling robust aggregation, while the client-side incorporates {Complementary Classification Fusion} (CCF) to balance generalization and personalization and {Cost-aware Feature Transmission} (CFT) for communication regulation.
    \item Through extensive experiments conducted across multiple heterogeneous scenarios and diverse classification datasets, we demonstrate FedMate's consistent effectiveness and stability. To validate its practical utility, we evaluate FedMate on semantic segmentation tasks using real-world autonomous driving datasets, which confirms its scalability and robustness in complex applications.
    %Extensive experiments across diverse heterogeneous scenarios and classification datasets demonstrate FedMate’s effectiveness and stability. Further evaluations on semantic segmentation tasks with autonomous driving datasets highlight its scalability to real-world applications.
\end{itemize}

\section{Related Work}
In this section, we critically analyze existing approaches from two fundamental perspectives: the partitioning strategies for personalized modules within local models, and the mechanisms for establishing and integrating generalization-oriented consensus. 
By discussing the inherent limitations of these methods, we elucidate the key motivations underlying our proposed approach.

\vspace{-7pt}
\subsection{Personalized Module Partitioning}
Traditional FL methods such as FedAvg~\cite{mcmahan2017communication} treat the entire model as a shared component and adopt identical training procedures across clients. MOON~\cite{li2021model} conducts contrastive learning within the feature space to drive local models away from their historical states and closer to the current global model. FedProx~\cite{li2020federated} and Ditto~\cite{li2021ditto}, on the other hand, introduce direct parameter-level constraints to regularize local training. 
Collectively, these methods attempt to utilize a unified global consensus to offer a general update direction for local models facing divergent objectives due to data heterogeneity~\cite{ye2023heterogeneous,wang2024comprehensive}. 
Nonetheless, these methods often enforce strong generalization by overwriting local models with the global model prior to training. Under substantial cross-client heterogeneity, selectively incorporating global consensus can better align with diverse client objectives.

Consequently, researchers have explored model partitioning strategies, selectively applying global updates to specific subsets of parameters~\cite{pillutla2022federated,deng2024fedasa}. This fine-grained approach seeks to enable the output model to more effectively balance global generalization with client-specific personalization. 
FedCAC~\cite{wu2023bold} conducts layer- and parameter-wise analysis by evaluating the degree of variation in model outputs or loss functions, generating masks to differentiate personalized layers from shared layers. In parallel, FedDecomp~\cite{wu2024decoupling} decomposes model parameters into the sum of personalized and shared components to achieve a clear separation of personalization and generalization knowledge. Notably, these methods typically introduce additional computational overhead during region partitioning or parameter decomposition. 
Motivated by the positive effects demonstrated in multi-task learning~\cite{bengio2013representation}, the strategy of decoupling the model into a feature extractor and a classifier has also garnered increasing attention in the PFL field. FedPer~\cite{arivazhagan2019federated} shares only the local feature extractor across clients, while FedRep~\cite{collins2021exploiting} builds upon FedPer by decoupling the training of the feature extractor and classifier. FedBABU~\cite{oh2022fedbabu} and Fed-RoD~\cite{chen2022on} further optimize the personalized classifier heads through local fine-tuning and vanilla softmax. However, these approaches offer limited consideration of the coupling between decoupled components and the internal balance of capacities within the classifier.

\vspace{-10pt}
\subsection{Generalization Consensus}
To achieve the personalization-aware generalization consensus, methods like FedPAC~\cite{xu2023personalized}, FedAMP~\cite{huang2021personalized}, and FedReMa~\cite{liang2024fedrema} leverage auxiliary local data statistics or model similarity measures to identify analogous clients for personalized aggregation. 
In addition, category center loss, widely adopted in FL for its representation learning efficacy~\cite{wen2016discriminative,florea2020margin, liu2024vmamba}, guides local training with generalization-oriented signals. Recent advances further employ global prototypes as a generalization consensus to regularize local feature learning:
% In terms of guiding local training with generalization-oriented information, category center loss has gained significant traction in FL due to its effectiveness in representation learning~\cite{wen2016discriminative, abramson2024accurate, liu2024vmamba}. 
% Building on this, recent works have explored the use of global prototypes as a generalization consensus to regularize the generation of local representations. 
FedPAC~\cite{xu2023personalized}, FedFA~\cite{zhou2023fedfa}, and FedProto~\cite{tan2022fedproto} use prototypes to regularize feature extractor training; FedProc~\cite{mu2023fedproc} applies them to contrastive learning; and FedPCL~\cite{tan2022federated} integrates them into personalized model learning.
Complementary approaches like FedMD~\cite{li2019fedmd} and FedKD~\cite{wu2022communication} distill global generalization knowledge into local models via classification-focused optimization. 
%employ knowledge distillation to optimize local classification decisions, thereby incorporating global generalization knowledge into personalized models.
Despite their promising capabilities, these methods often lack robust mechanisms to rigorously assess the quality of uploaded information, which undermines the reliability of the global consensus.

\begin{figure*}[h]
    \centering
    \includegraphics[width=0.83\textwidth]{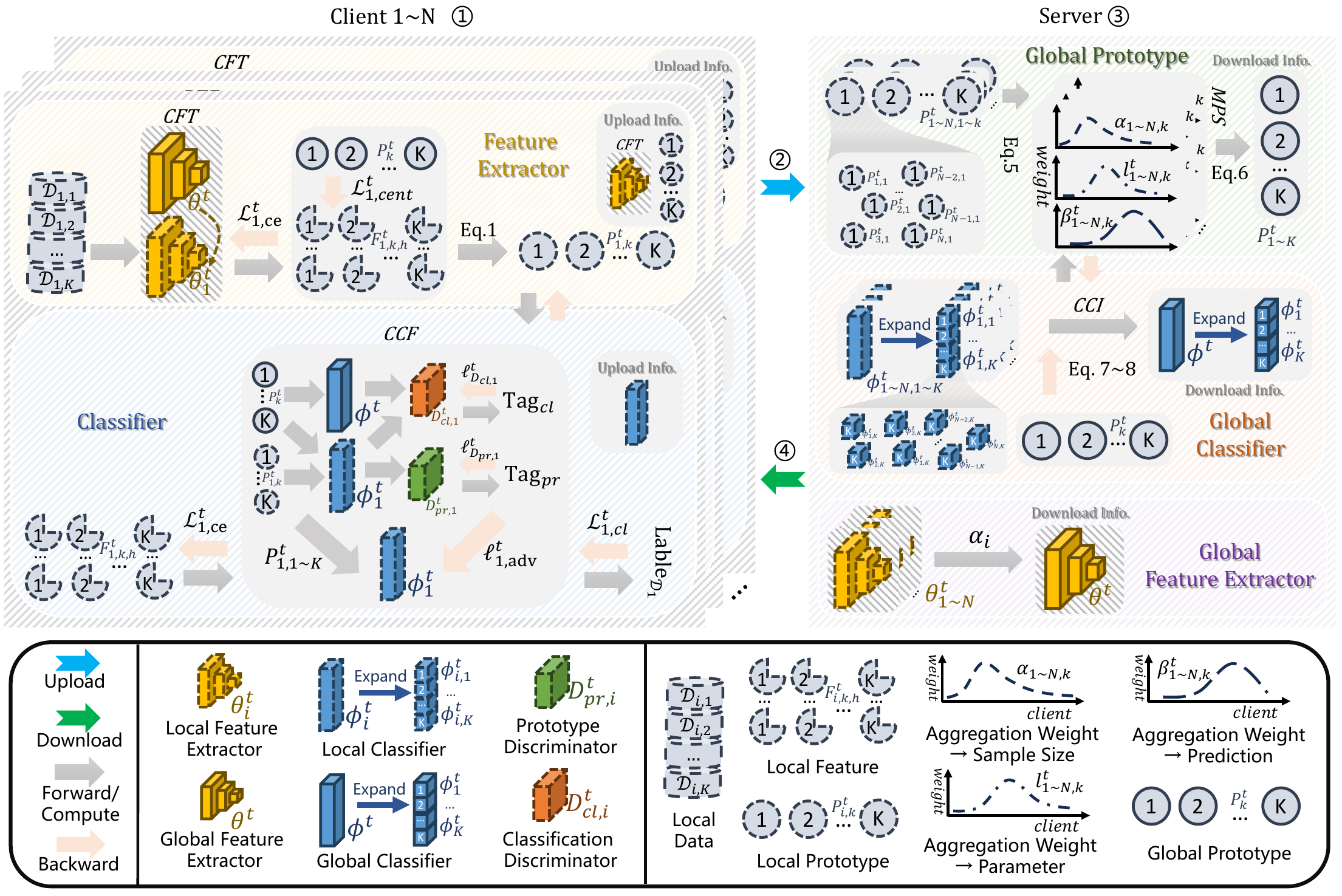}
    \vspace{-5pt}
    \caption{FedMate workflow. \ding{172} Local training: Clients train with decoupled objectives (cross-entropy loss, center loss, and dual adversarial learning). \ding{173} Selective uploading: Clients upload local classifiers, prototypes, and sample statistics (both total and per-class counts); feature extractors are uploaded only during scheduled rounds. \ding{174} Server aggregation: The server weights multi-view prototypes using relative entropy-based scoring and aggregates classifiers per-class via sample-size-weighted averaging with fine-tuning. \ding{175} Global broadcast: The server distributes updated global classifiers and prototypes to all clients; the global feature extractor is downloaded selectively.}
    \label{flow}
    \vspace{-5pt}
\end{figure*}

\vspace{-2pt}
\section{Methodology}
\vspace{-1pt}
\subsection{Preliminaries and Problem Formulation}
\vspace{-2pt}
\textbf{Terminology.} We consider an FL scenario with $N$ clients, where the union of all local datasets is denoted as $\mathcal{D}$. The dataset owned by client $i$ is denoted by $\mathcal{D}_i$, satisfying $\mathcal{D} = \bigcup_{i=1}^{N} \mathcal{D}_i$. The total dataset $\mathcal{D}$ spans a set of classes $\mathcal{C}$, where $|\mathcal{C}|$ denotes the total number of classes, and class indices are stored in $[\mathcal{C}]$. 
Each client $i$ holds a subset of classes denoted as $[\mathcal{C}_i] \subseteq [\mathcal{C}]$. Each sample $(x_i, y_i) \in \mathcal{D}_i$ consists of an input $x_i \in \mathcal{X}$ and a label $y_i \in [\mathcal{C}_i]$, where the input space satisfies $\mathcal{X} \subseteq \mathbb{R}^d$ with $d$ denoting the input dimension. The model of client $i$ is parameterized by $w_i$, consisting of a feature extractor $\theta_i$ and a classifier $\phi_i$. The feature extractor $f_i: \mathbb{R}^d \to \mathbb{R}^K$ maps inputs from $\mathcal{X}$ to a representation space $\mathcal{H} \subseteq \mathbb{R}^K$. 
For each class $k \in [\mathcal{C}_i]$, the class prototype $P_{i,k}$ is computed as:
\begin{equation}\label{eq1}
    P_{i,k} = \frac{\sum_{x_{i,k} \in \mathcal{D}_{i,k}} f_i(\theta_i; x_{i,k})}{|\mathcal{D}_{i,k}|},
    \vspace{-2pt}
\end{equation}
where $\mathcal{D}_{i,k}$ is the set of local samples from class $k$. The classifier $g_i: \mathbb{R}^K \to \mathbb{R}^{|\mathcal{C}|}$ maps feature embeddings to logits  over the full label space $\mathcal{C}$, with final predictions obtained by normalizing the logits. 
%The logits are further normalized (e.g., via softmax) to produce the final prediction. 
Throughout this paper, the final layer is defined as the classifier, and all preceding layers collectively form the feature extractor.

\textbf{Loss function and data heterogeneity.} 
% This paper focuses on PFL, where each client maintains a personalized model optimized for its local task. 
This paper studies PFL, where each client retains a model tailored to its local task.
%To achieve this, each client maintains a personalized model and computes the loss based on its local training data. 
The system objective minimizes the weighted average of client-specific losses:
\begin{equation}\label{eq2}
\mathcal{L}(W) = \sum_{i=1}^{N} \alpha_i \frac{\sum_{(x_i, y_i) \in \mathcal{D}_i} \ell_{\mathrm{ce}}(\omega_i; x_i, y_i)}{|\mathcal{D}_i|}, \quad \alpha_i = \frac{|\mathcal{D}_i|}{\sum_{j=1}^N |\mathcal{D}_j|},
\vspace{-2pt}
\end{equation}
where $W = \{\omega_1, \omega_2, \dots, \omega_N\}$ is the set of all client models, $\ell_{\mathrm{ce}}(\cdot)$ denotes the cross-entropy loss, and $\alpha_i$ is the normalized weight proportional to the normalized $|\mathcal{D}_i|$. For the $i$-th client, the joint data distribution $\mathbb{P}(x_i, y_i)$ decomposes into: $\mathbb{P}(x_i, y_i) = \mathbb{P}(x_i|y_i) \mathbb{P}(y_i) = \mathbb{P}(y_i|x_i) \mathbb{P}(x_i)$, where $\mathbb{P}(x_i)$ and $\mathbb{P}(y_i)$ are the marginal distributions over the input and label spaces, respectively.
We specifically address client-wise label distribution skew, where different clients have distinct marginal label distributions, formally expressed as $\mathbb{P}(y_i) \neq \mathbb{P}(y_j)$ for $i \neq j$. This heterogeneity challenges the learning of robust and generalizable models.
\vspace{-5pt}
\subsection{FedMate Overview}
\textbf{Step 1:} \textit{Client training}. 
(i) Each client performs supervised training with cross-entropy loss on its local dataset, following a decoupled strategy similar to FedRep~\cite{collins2021exploiting}, where the classifier is trained before the feature extractor.
(ii) The feature extractor training incorporates a center loss computed using global prototypes to enhance feature compactness.
(iii) The classifier training employs an adversarial loss with dual discriminators to align global and local prototype distributions, treating the classifier as a generator, and a prototype-based cross-entropy loss to maintain knowledge of previous classes and mitigate catastrophic forgetting. 

\textbf{Step 2:} \textit{Client-to-server upload}. Clients upload their local classifier, prototypes, and both total and per-class sample sizes. The local feature extractor is selectively uploaded only when the current round belongs to the predefined subset of rounds $Q$.

\textbf{Step 3:} \textit{Server aggregation}.
(i) For prototypes, the weights used in aggregation are computed from sample sizes, prototype parameters, and predictions of the previous global classifier, then combined via relative entropy. 
(ii) For classifiers, a fine-grained per-class aggregation is performed, weighted by per-class sample sizes. The aggregated classifier is fine-tuned using the latest global prototypes. 
(iii) If feature extractor parameters are received, they are aggregated weighted by the total sample sizes.

\textbf{Step 4:} \textit{Server-to-client download}. The server broadcasts the global classifier and prototypes to all clients, while the global feature extractor is only transmitted to clients when the current communication round involves feature extractor aggregation, maintaining synchronization with the conditional upload protocol from Step~2.
%is selectively downloaded only if local feature extractors were aggregated in the current round.

This process repeats until convergence or the maximum communication round is reached (see Fig.~\ref{flow} and Appx.~1 for details).

\vspace{-5pt}
\subsection{Complementary Classification Fusion (CCF)}
To effectively fuse complementary classification knowledge, we first differentiate the strengths of global and local classifiers. The global classifier provides balanced knowledge coverage and improved per-class generalization by aggregating cross-client information. In contrast, the local classifier specializes in task-specific adaptation through sustained training on local datasets. Our goal is to leverage the combined knowledge of these classifiers to (i) fill class-specific gaps in the local classifier, (ii) enhance its generalization over existing classes, and (iii) mitigate catastrophic forgetting of local adaptability. 
Since the global prototype is aggregated from client-specific local prototypes, it inherently reflects the distribution of global tasks, whereas each local prototype captures the task-specific characteristics of its respective client.
To retain valuable information from both perspectives, we introduce an adversarial training mechanism that selectively preserves the most beneficial classification knowledge.

Specifically, our approach implements two distinct adversarial training objectives. First, at round $t$, the global prototypes 
$P^t = \{ P^t_1, P^t_2, \dots, P^t_{|\mathcal{C}|}\}$ and the local prototypes $P^t_i = \{ P^t_{i,k} \mid k \in [\mathcal{C}_i] \}$ are input into the local classifier $\phi_i^t$. The resulting outputs are then passed to the prototype discriminator $D_{pr,i}$ (a binary discriminator producing probability-valued outputs) trained with loss: 
\begin{align*}
\ell_{D_{pr,i}}^t = \sum_{k\in[\mathcal{C}_i]} \Big[ 
    & \log D_{pr,i}\big(g_i(\phi_i^t; P^t_k)\big) \\
  +\; & \log\big( 1 - D_{pr,i}(g_i(\phi_i^t; P_{i,k}^t)) \big) 
\Big].
\end{align*} 
In the second phase, the global prototypes $P^t$ are evaluated using both the local ($\phi_i^t$) and global ($\phi^t$) classifiers. The outputs are then assessed by a classification discriminator $D_{cl,i}$ (similarly a binary discriminator yielding probabilistic scores), which minimizes:
% \begin{equation*}
% \ell_{D_{cl,i}}^t = \sum_{k\in[\mathcal{C}]} \Big[ \log D_{cl,i}\big(g(\phi^t; P^t_k)\big) + \log\big( 1 - D_{cl,i}(g_i(\phi^t_i; P^t_k)) \big) \Big].
% \end{equation*}
\begin{align*}
\ell_{D_{cl,i}}^t = \sum_{k\in[\mathcal{C}]} \Big[
& \log D_{cl,i}\big(g(\phi^t; P^t_k)\big) \\
+\; & \log\big( 1 - D_{cl,i}(g_i(\phi^t_i; P^t_k)) \big)\Big].
\end{align*}
\begin{linenomath}
Accordingly, the local classifier \( \phi_i^t \) serves as the generator in both adversarial processes, guided by feedback from the discriminators:
$$\ell_{i,\text{adv}}^t = \sum_{k\in[\mathcal{C}_i]} \log D_{pr,i}(g_i(\phi_i^t; P_{i,k}^t)) + \sum_{k\in[\mathcal{C}]} \log D_{cl,i}(g_i(\phi_i^t; P_k^t)),
$$
where \( \ell_{i,\text{adv}}^t \) is the adversarial loss. To further alleviate forgetting, the local prototype $P^t_i$ is introduced as an additional supervisory signal: 
$$
\ell_{i,\text{ccf}}^t = \sum_{k\in[\mathcal{C}_i]}\ell_{\text{ce}}(\phi_i^t; P_{i,k}^t,y_{i,k}) + \kappa_{cl}^t \ell_{i,\text{adv}}^t, \quad \kappa_{cl}^t = 1 - \frac{t}{t_{\max}},
$$
\end{linenomath}
where \( \ell_{i,\text{ccf}}^t \) is the CCF component loss, and \( \kappa_{cl}^t \) is a weighting factor that balances the adversarial loss. For stabilize early training, \( \kappa_{cl}^t \) is adjusted based on the current round \( t \) and the maximum round \( t_{\text{max}} \). The overall loss for training the local classifier is then given by:
\begin{equation}\label{eq3}
    \ell_{i,cl}^t = \sum_{(x_i,y_i)\in \mathcal{D}_i}\ell_{\mathrm{ce}}(\omega_i^t; x_i, y_i) + \lambda_c\ell_{i,\text{ccf}}^t,
\end{equation}
where the hyperparameter \( \lambda_c \) controls the contribution of $\ell_{i,\text{ccf}}^t$.
A detailed description of the CCF input–output workflow is in Appx.~3.
% The regularization term adjustment is controlled by .

% \vspace{-10pt}
\subsection{Cost-aware Feature Transmission (CFT)}
%FedProto~\cite{tan2022fedproto} transmits only prototypes rather than feature extractors. However, 
When the global prototypes possess constrained representational capacity (e.g., due to low-dimensional embeddings), their effectiveness in guiding local feature extractor training becomes limited. Prior solutions like FedPAC~\cite{xu2023personalized} and FedFA~\cite{zhou2023fedfa} attempt to mitigate this by transmitting both global feature extractors and prototypes to clients, forcibly overwriting local extractors before prototype-based supervision. While this strategy improves global model consistency, it introduces substantially increased communication overhead from frequent extractor transfers and potential degradation of local model personalization due to overwriting of client-specific features.
%increase communication costs and may compromise local personalization.

Motivated by these limitations, we propose a cost-aware selective communication strategy for feature extractor transmission. Our approach dynamically schedules transmissions based on the information ratio between prototypes and the feature extractor, thereby reducing unnecessary communication overhead while balancing the generalization–personalization trade-off for local models. To enhance deployability, we quantify information content using parameter counts, defining the information ratio $q$ between prototypes and feature extractor as:
\begin{equation}\label{eq4}
q = \frac{Par\left(\theta_i^t\right)}{Par\left(\sum_{k\in[\mathcal{C}]}P_{k}^t\right)},
\end{equation}
where $Par(\cdot)$ denotes the parameter count function. The feature extractor is selectively uploaded at rounds that are not multiples of $x \times q$, maintaining equivalent total communication costs to per-round model uploading. In this paper, we collect the selected rounds into a scheduling set $Q$. 

\subsection{Multi-view Prototype Scrutiny (MPS)}
As discussed in Section~\ref{intro}, determining prototype aggregation weights based solely on local sample sizes may yield suboptimal global representations. To address this, we propose a multi-faceted weight calibration approach that jointly considers: sample sizes, prototype parameters, and predictions from the previous global classifier. This integrated weighting scheme promotes a more robust and unbiased global consensus during prototype aggregation. 

For each class \( k\in [\mathcal{C}_i]\), we compute three complementary weighting components: (i) sample-size-based \( \alpha_{i,k} \), determined by the sample size \( |\mathcal{D}_{i,k}| \); (ii) centroid-similarity-based \( l_{i,k}^t \), obtained by the cosine similarity \( sim \) between each local prototype and its parameter centroid \( An_k^t \); and (iii) prediction-based \( \beta_{i,k}^t \), derived from logit-based evaluation of each prototype using the previous global classifier \( \phi^{t-1} \). These weights are defined as: 
\begin{equation*}
\alpha_{i,k} = \frac{|\mathcal{D}_{i,k}|}{\sum_{j\in \mathcal{B}_k} |\mathcal{D}_{j,k}|}, 
\end{equation*}
\begin{nolinenumbers}
\begin{equation*}
l_{i,k}^t = \frac{sim(P_{i,k}^t, An_k^t)}{\sum_{j\in \mathcal{B}_k} sim(P_{j,k}^t, An_k^t)},
\end{equation*}
\begin{equation}\label{eq5}
\beta_{i,k}^t = \frac{g(\phi^{t-1}_k, P_{i,k}^t)}{\sum_{j\in \mathcal{B}_k} g(\phi^{t-1}_k, P_{j,k}^t)}, 
\end{equation}
\end{nolinenumbers}
where $An_k^t = \frac{\sum_{i\in \mathcal{B}_k} P_{i,k}^t}{|\mathcal{B}_k|} $ with \( \mathcal{B}_k \) denoting the set of clients with local prototypes for class \( k \), and $\phi_{k}^{t-1}$ is the $k$-th class neuron parameters of global classifier. 

Then, we compute pairwise Jensen-Shannon (JS) divergences between the three weight sets $\mathcal{S}_{i,k}^t = \{\alpha_{i,k},\ l_{i,k}^t,\ \beta_{i,k}^t\}$, requiring only three comparisons due to JS symmetry. These divergences are aggregated into a consensus weighting term \( W_\text{mps} \): 
\begin{equation*}
W_{i,k,\text{mps}}^t = \text{Softmax}\left( -\left\{ \sum_{s' \neq s} \text{JS}(s, s')\ |\ s \in \mathcal{S}_{i,k}^t \right\} \right).
\end{equation*}
Smaller JS divergence yields larger normalized weights, penalizing deviant components. 
The final aggregation weight combines all three components: 
\begin{equation*}
W_{i,k,\text{final}}^t = W_{i,k,\text{mps}_1}^t \cdot \alpha_{i,k} + W_{i,k,\text{mps}_2}^t \cdot l_{i,k}^t + W_{i,k,\text{mps}_3}^t \cdot \beta_{i,k}^t,
\end{equation*}
where $W_{i,k,\text{mps}_h}^t$ denotes the $h$-th element of $W_{i,k,\text{mps}}^t$. Now we obtain the aggregated global prototype as:
%Subsequently, $W_{\text{mps}}$ is used to reweight $\alpha$, $l$, and $\beta$, producing the final aggregated weight $W_{\text{final}}$ for prototype aggregation:
\begin{equation}\label{eq6}
P^t_k=\sum_{i\in\mathcal{B}_k}W_{i,k,\text{final}}^t\cdot P_{i,k}^t.
\end{equation}

\subsection{Category-wise Classifier Integration (CCI)}
In classification architectures, output neurons in the classifier is inherently linked to a specific class, serving as a direct predictor for that category. Conventional aggregation methods typically process the classifier holistically, disregarding this intrinsic neuron-class alignment. Such coarse-grained fusion risks introducing bias toward client-specific dominant categories, particularly under heterogeneous data conditions where class samples are unevenly distributed across clients. 
To address this, we propose the CCI scheme, which performs neuron-level aggregation aligned with class boundaries. This fine-grained approach enables more targeted and unbiased parameter fusion, reducing the impact of client-side data heterogeneity and refining each category's decision boundary based on available knowledge.

For each class $k$, we aggregate the corresponding classifier neuron parameters across clients by sample-size-proportional weighting: 
\begin{equation}\label{eq7}
\phi_k^t = \alpha_{i,k} \cdot \phi_{i,k}^t.
\end{equation}
% where $\phi_{i,k}^t$ is the $k$-th class neuron parameters of client $i$'s classifier. 

Furthermore, to harmonize the decoupled training of feature extractors and classifiers, we enforce consistency between the global prototypes (class-level feature centroids) and the global classifier. While prototypes guide feature extractor learning and classifier parameters direct local optimization, their alignment ensures coherent joint evolution of the feature-classification pipeline. This prevents objective drift between modules during the FL process. Specifically, the global classifier is fine-tuned using the current global prototypes:
%The global prototypes, as , guide the learning of the feature extractor, while the global classifier parameters guide local classifier optimization. Ensuring alignment between these two allows feature extraction and classification to evolve coherently during training, preventing objective drift between modules. In practice, the global classifier is fine-tuned using the latest global prototypes: 
\begin{equation}\label{eq8}
\phi_k^t \leftarrow \phi_k^t - \eta \nabla_{\phi} \ell_{\text{ce}} \left( \phi_k^t; P_k^t,y_k \right),    
\end{equation}
where $\eta$ is the fine-tuning learning rate. This prototype-aware refinement aligns decision boundaries with evolving feature clusters, maintaining coupling between the feature extractor and classifier despite local decoupled training.

\section{Experiments}
In this section, we present key experimental results evaluating the generalization, adaptability, stability, and scalability of our method. Additional analyses of convergence, communication costs, and hyperparameter sensitivity are provided in Appx.~2.
% In this section, we conduct extensive experiments to evaluate the effectiveness of our method across multiple dimensions. We validate its generalization and local adaptability on diverse benchmarks, and provide feature-space visualizations to illustrate training stability. Results on semantic segmentation further demonstrate its scalability to more complex tasks. Additional results on convergence and further analyses are provided in Appx.~\ref{appb}.
\subsection{Experimental Setup}
\textbf{Datasets.} 
We evaluate our method's generalization and robustness across six benchmark datasets spanning varying complexity levels: CIFAR-10~\cite{krizhevsky2009learning}, CINIC-10~\cite{darlow2018cinic}, Animal-10~\cite{corrado2018animals10}, EMNIST~\cite{cohen2017emnist}, CIFAR-100~\cite{krizhevsky2009learning}, and Cityscapes~\cite{cordts2016cityscapes}. These datasets differ in input dimensionality, label granularity, and visual complexity, offering a comprehensive foundation for performance evaluation.

\textbf{Implementation details.}
We employ dataset-specific architectures with modular feature extractors and classifiers: CIFAR-10 and CINIC-10 use a 3-Conv-2-FC CNN (following FedPAC~\cite{xu2023personalized}); Animal-10 adopts an enhanced 3-Conv-2-FC variant with larger filters for natural images; EMNIST utilizes a compact 2-Conv design; CIFAR-100 implements a deeper 4-Conv network; and Cityscapes employs BiSeNetV2 for segmentation,  demonstrating scalability to deeper, dense prediction tasks. Additionally, following established protocols from SCAFFOLD~\cite{karimireddy2020scaffold} and FedAMP~\cite{huang2021personalized}, we construct client datasets with controlled class imbalance through a tunable parameter \( s \in [0,100]\): \( s\% \) of each client's training data follows uniform class distribution, while the remaining \( (100 - s)\% \) exhibits bias toward randomly selected dominant classes. Crucially, all clients share an identically structured test set with balanced class distribution and equivalent total samples (matching training set size), ensuring evaluation reflects true generalization across the complete label space rather than local data biases. This design mimics realistic scenarios where future data distributions are unknown.
%thus introducing class distribution imbalance. At the same time, we assign each client a test set with the same number of samples per class and the same total number of samples. This provides a neutral evaluation baseline that is not affected by class distribution (since the future data environment is unpredictable), in order to fairly evaluate the model's generalization ability across the entire label space. 

%For image classification, we adopt lightweight CNNs tailored to each dataset; CIFAR-10 and CINIC-10 use a 3-Conv-2-FC architecture following FedPAC; Animal-10 shares the same depth with adjustments for natural image complexity; EMNIST uses a shallower 2-Conv model due to its lower task difficulty; CIFAR-100 adopts a deeper 4-Conv variant; and for Cityscapes segmentation, we use BiSeNetV2 to demonstrate scalability to deeper, dense prediction tasks. All settings modularize the feature extractor and classifier. 

\textbf{Comparison baselines.}
We evaluate against three categories of FL approaches: 1) traditional FL methods, including  FT-FedAvg~\cite{mcmahan2017communication} (post-training fine-tuning) and MOON~\cite{li2021model} (contrastive representation alignment); 2) prototype-based methods, such as FedFA~\cite{zhou2023fedfa} (joint extractor-classifier regularization) and FedProto~\cite{tan2022fedproto} (prototype-only aggregation); and 3) model-decoupling approaches, comprising LG-FedAvg~\cite{liang2020think} (personalized extractors), FedPer~\cite{arivazhagan2019federated}/FedRep~\cite{collins2021exploiting} (personalized classifiers with shared extractors, where FedRep additionally decouples training phases), FedBABU~\cite{oh2022fedbabu} (phased classifier freezing), and FedPAC~\cite{xu2023personalized} (statistics-augmented personalization). For segmentation tasks, we include FedSeg~\cite{miao2023fedseg} (contrastive-enhanced representation learning). All classification experiments report mean accuracy ($\%$).
%performs local fine-tuning on the global model after conventional federated training, while MOON~\cite{li2021model} incorporates representation-level contrastive learning to align local and global models. FedFA~\cite{zhou2023fedfa} leverages globally aggregated prototypes to jointly regularize the local extractor and classifier, whereas FedProto~\cite{tan2022fedproto} aggregates only class prototypes, which serve as supervision information for local training.
%In terms of model decoupling, LG-FedAvg~\cite{liang2020think} personalizes the feature extractor while sharing a global classifier; conversely, FedPer~\cite{arivazhagan2019federated} and FedRep~\cite{collins2021exploiting} share the extractor and personalize the classifier, with FedRep further decoupling their training processes—a strategy also adopted in our work. FedBABU~\cite{oh2022fedbabu} freezes the classifier during extractor training and performs local fine-tuning post-optimization. FedPAC~\cite{xu2023personalized} uploads auxiliary statistics to support personalized classifier aggregation. For segmentation, FedSeg~\cite{miao2023fedseg} enhances models with cross-entropy loss optimization and contrastive learning in the representation space.
%The following experiments use mean accuracy ($\%$) for evaluation.
\subsection{Main Results}
\begin{table*}[h!]
\centering
\resizebox{\textwidth}{!}{
\begin{tabular}{lcccccccc}
\toprule
\multirow{2}{*}{\textbf{Method}} & \multicolumn{4}{c}{\textbf{CIFAR-10}} & \multicolumn{2}{c}{\textbf{CINIC-10}} & \multicolumn{2}{c}{\textbf{EMNIST}} \\ \cmidrule(lr){2-5} \cmidrule(lr){6-7} \cmidrule(lr){8-9} 
                                   & \textbf{s = 10} & \textbf{s = 30} & \textbf{s = 50} & \textbf{s = 70} & \textbf{s = 10} & \textbf{s = 70} & \textbf{s = 10} & \textbf{s = 70} \\ \midrule
\textbf{Local}                    & 28.61$\pm$0.06   & 36.08$\pm$0.07   & 40.51$\pm$0.05 & 42.40$\pm$0.04 & 22.86$\pm$0.03 & 29.95$\pm$0.04 & 31.02
$\pm$0.09 & 50.84
$\pm$0.08 \\ \midrule
\textbf{FT-FedAvg}                   & 49.39$\pm$0.07   & 57.55$\pm$0.09   & 61.95$\pm$0.08 & 64.33$\pm$0.06 & 32.72$\pm$0.06 & 40.55$\pm$0.07 & 59.03
$\pm$0.11 & 69.24
$\pm$0.05 \\ \midrule
\textbf{FedProto}                    & 29.91$\pm$0.04   & 38.45$\pm$0.07   & 43.54$\pm$0.08 & 46.27$\pm$0.10 & 22.19$\pm$0.02 & 30.15$\pm$0.04 & 31.72
$\pm$0.09 & 57.14
$\pm$0.06 \\ \midrule
\textbf{LG-FedAvg}                     & 28.32$\pm$0.05   & 35.79$\pm$0.06   & 40.91$\pm$0.04 & 42.37$\pm$0.11 & 22.57$\pm$0.04 & 29.14$\pm$0.03 & 31.71
$\pm$0.05 & 53.64
$\pm$0.04 \\ \midrule
\textbf{FedPer}                     & 33.93$\pm$0.06   & 42.39$\pm$0.05   & 47.49$\pm$0.07 & 51.43$\pm$0.10 & 24.64$\pm$0.05 & 32.60$\pm$0.03 & 36.39
$\pm$0.06 & 54.97
$\pm$0.07 \\ \midrule
\textbf{FedRep}                     & 38.97$\pm$0.09   & 47.63$\pm$0.10   & 52.53$\pm$0.05 & 53.51$\pm$0.06 & 26.64$\pm$0.03 & 33.92$\pm$0.04 & 34.08
$\pm$0.09 & 55.47
$\pm$0.08 \\ \midrule
\textbf{FedBABU}                     & 45.05$\pm$0.08   & 56.25$\pm$0.07   & 60.89$\pm$0.06 & 64.87$\pm$0.05 & 27.45$\pm$0.12 & 37.12$\pm$0.03 & 30.69
$\pm$0.08 & 46.23
$\pm$0.06 \\ \midrule
\textbf{FedPAC}                   & 53.02$\pm$0.06   & 65.42$\pm$0.08   & 69.97$\pm$0.07 & 71.08$\pm$0.04 & \textbf{35.67$\pm$0.05} & 49.09$\pm$0.07 & 56.71
$\pm$0.09 & 70.14
$\pm$0.06 \\ \midrule
\textbf{Ours}                       & \textbf{55.05$\pm$0.02}       & \textbf{68.13$\pm$0.11}       & \textbf{70.21$\pm$0.02}     & \textbf{72.19$\pm$0.06}     & {34.98$\pm$0.02}     & \textbf{50.24$\pm$0.06}     & \textbf{60.29$\pm$0.03}     & \textbf{72.15
$\pm$0.10}     \\
           % & \textcolor{green}{↑ 2.02} & \textcolor{green}{↑ 2.71} & \textcolor{green}{↑ 0.24} & \textcolor{green}{↑ 1.11} & \text{-} & \textcolor{green}{↑ 1.15} & \textcolor{green}{↑ 1.26} & \textcolor{green}{↑ 2.01} \\ 
\midrule
\textbf{Backbone}                       & 39.11 $\pm$0.05   & 48.17$\pm$0.10   & 52.51$\pm$0.06 & 54.20$\pm$0.08 & 26.33$\pm$0.11 & 33.95$\pm$0.02 & 35.47
$\pm$0.05 & 55.37
$\pm$0.08     \\
\textbf{+\textit{MPS}+\textit{CFT}}                       & 50.17$\pm$0.10   & 59.22$\pm$0.05   & 63.26$\pm$0.08 & 66.75$\pm$0.06 & 29.66$\pm$0.03 & 45.07$\pm$0.05 & 52.49
$\pm$0.05 & 67.33
$\pm$0.07     \\
\textbf{+\textit{CCI}+\textit{CCF}}                       & 52.17$\pm$0.03   & 62.15$\pm$0.09   & 62.89$\pm$0.12 & 64.97$\pm$0.03 & 30.02$\pm$0.08 & 44.27$\pm$0.08 & 54.17
$\pm$0.11 & 66.57
$\pm$0.05     \\
           \bottomrule
\end{tabular}
}
\caption{Generalization performance evaluation under varying heterogeneity settings (controlled by $s$).}
\label{tab:test-generalization}
\end{table*}

\textbf{Generalization performance evaluation.} Our comprehensive experiments assess model generalization under varying data heterogeneity levels (Tab.~\ref{tab:test-generalization}). Key findings reveal: 1) our method consistently outperforms the baselines across datasets and heterogeneity scenarios; 2) standard model-decoupling PFL approaches (e.g., LG-FedAvg, FedPer, and FedRep) show limited effectiveness in global complementary knowledge integration; 3) prototype-only methods (FedProto) demonstrate the inherent limitations of low-dimensional guidance (suffering a >15\% performance degradation compared to our approach); 4) our CCF mechanism enforces moderate global constraints during local training, enabling more effective knowledge integration and reducing overfitting.  
%Methods such as LG-FedAvg, FedPer, and FedRep fail to effectively integrate global complementary knowledge, which limits their performance. 
%In contrast, our method applies moderate global constraints through CCF during local training, enabling more effective knowledge integration and reducing overfitting. The limited performance of FedProto highlights the insufficiency of low-dimensional prototypes in guiding local training. Our approach, leveraging communication-aware selective transmission of feature extractors, subtly balances effectiveness with communication cost.

% \vspace{-10pt}
\textbf{Ablation analysis.} We systematically evaluate component contributions under identical heterogeneous conditions. The results are shown in the lower half of Tab.~\ref{tab:test-generalization}, which demonstrates that both feature extractor modules ({MPS} and {CFT}) and classifier components ({CCI} and {CCF}) contribute significantly to model performance, with their combined removal degrading accuracy by {3\%-17\% across datasets}. 
%{MPS} and {CFT} components in the feature extractor, and the {CCI} and {CCF} components in the classifier, as a cohesive unit due to their mutual reinforcement. The results demonstrate that all components are effective and that their synergy plays a crucial role in the performance of the model. 
Notably, under extreme heterogeneity ($s\leq 10$), introducing moderate global constraints at the classifier level substantially improves generalization, aligning with our objective stated in Section~\ref{intro}: achieving a balance between global generalization and local personalization through structurally decoupled yet functionally coordinated modules.

\textbf{Local adaptability evaluation.}
\begin{table}[!t]
\centering
\resizebox{0.47\textwidth}{!}{ 
\begin{tabular}{lccc}
\toprule
\textbf{Method} & \textbf{CIFAR-10} & \textbf{CINIC-10} & \textbf{CIFAR-100} \\ 
\midrule
\textbf{Local}  & 65.15$\pm$0.08 & 55.21$\pm$0.12 & 35.27$\pm$0.14 \\ \midrule
\textbf{FT-FedAvg}  & 90.65$\pm$0.07 & 84.12$\pm$0.03 & 70.51$\pm$0.10 \\ \midrule
\textbf{MOON} & 66.67$\pm$0.12 & 57.37$\pm$0.06 & 37.29$\pm$0.09 \\ \midrule
\textbf{FedProto} & 85.21$\pm$0.04 & 81.61$\pm$0.11 & 75.28$\pm$0.06 \\ \midrule
\textbf{FedFA} & 84.19$\pm$0.13 & 80.86$\pm$0.09 & 71.59$\pm$0.07 \\ \midrule
\textbf{LG-FedAvg} & 88.35$\pm$0.05 & 82.10$\pm$0.06 & 70.95$\pm$0.12 \\ \midrule
\textbf{FedPer} & 91.21$\pm$0.09 & 85.05$\pm$0.07 & 69.97$\pm$0.10 \\ \midrule
\textbf{FedRep} & 91.45$\pm$0.02 & 85.88$\pm$0.08 & 75.33$\pm$0.11 \\ \midrule
\textbf{FedBABU} & 88.83$\pm$0.10 & 81.65$\pm$0.13 & 67.71$\pm$0.12 \\ \midrule
\textbf{FedPAC} & 91.71$\pm$0.06 & 84.28$\pm$0.09 & 74.24$\pm$0.08 \\ \midrule
\textbf{Ours} & \textbf{92.18$\pm$0.13} & \textbf{86.01$\pm$0.10} & \textbf{77.04$\pm$0.05} \\
\bottomrule
\end{tabular}
}
\caption{Adaptability evaluation under extreme heterogeneity settings.}
\label{tab:method_comparison}
\vspace{-8pt}
\end{table}
To evaluate the models' adaptability to local tasks, we consider an extreme pathological heterogeneity scenario where each client holds data from a subset of classes (3 classes for CIFAR-10 and CINIC-10, 10 classes for CIFAR-100), with test sets matching local data distributions. Our method demonstrates superior adaptability across datasets of varying difficulty (Tab.~\ref{tab:method_comparison}) through two key design aspects: 1) selective integration of global feature extractors (avoiding full overwrites) preserves local feature specialization while enabling controlled knowledge transfer; and 2) prototype-augmented classifier guidance (combining global classifiers with local prototypes) prevents catastrophic forgetting of personalized knowledge. This dual mechanism achieves a {performance gain close to 2\% over baselines} on the challenging CIFAR-100 dataset while maintaining the critical generalization-personalization balance. 
%Experimental results, as shown in Tab.~\ref{tab:method_comparison}, indicate that our method performs competitively across datasets of varying difficulty. This is attributed to two aspects of our global constraint design: first, by not fully overriding the local feature extractor with the global one in every round, we enable moderate integration; second, when using the global classifier to guide local training, incorporating local prototypes helps mitigate the forgetting of personalized knowledge. This design strikes a balance between generalization and personalization. 
%Additionally, we find that fine-tuning improves performance—FedAvg, when followed by local fine-tuning after each round, leads to notable performance gains.
\begin{figure}[t]
    \centering
    \includegraphics[width=0.35\textwidth]{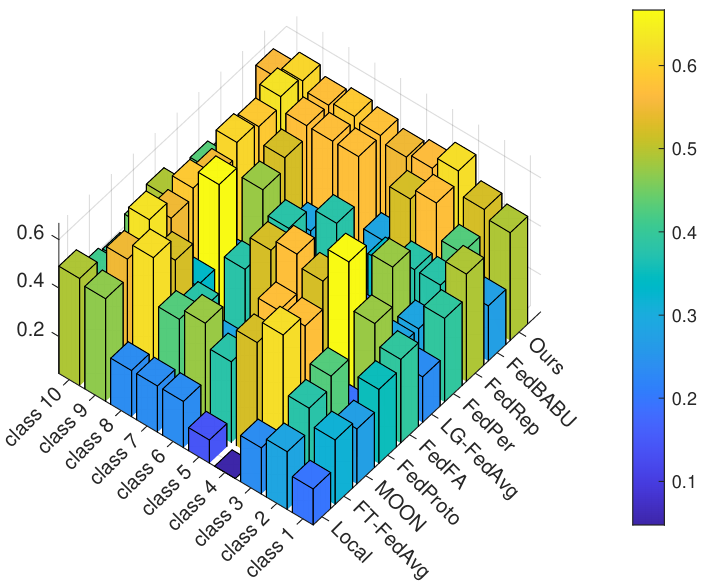}
    \caption{Class-wise accuracy on Animal-10.}
    \label{fig:cl-ac}
\end{figure}

\textbf{Class-wise accuracy analysis.} 
We evaluate per-class performance on the {Animal-10} dataset (high inter-class similarity) using personalized models trained under data heterogeneity \(s = 20\), with uniformly distributed test sets. As shown in Fig.~\ref{fig:cl-ac}, standard methods like {FedRep} and {FedPer} exhibit strong performance on locally abundant classes but fail on underrepresented ones, reflecting their inability to correct for biased client uploads during global aggregation. Our approach dynamically adjusts aggregation weights using MPS and incorporates class-wise fine-grained aggregation for the classifier, achieving balanced accuracy across all classes.
%where classes are highly similar in feature space, providing a clear test of the model’s ability to capture class-specific information. The training dataset has a heterogeneity degree of \(s = 20\), and the test set is uniformly distributed. As shown in Fig.~\ref{fig:cl-ac}, methods like {FedRep} and {FedPer} perform well on classes with sufficient local samples but degrade on underrepresented classes due to their inability to account for biases in the uploaded information, leading to a skewed global consensus and incomplete knowledge integration.
%Conversely, our method dynamically adjusts aggregation weights using MPS and incorporates class-wise fine-grained aggregation at the classifier, which improves adaptability and balances performance across classes.

\textbf{Stability and robustness analysis.}
\begin{figure}[!t]
    \centering
    \subfigure[FedFA]{
    \includegraphics[width=0.23\textwidth]{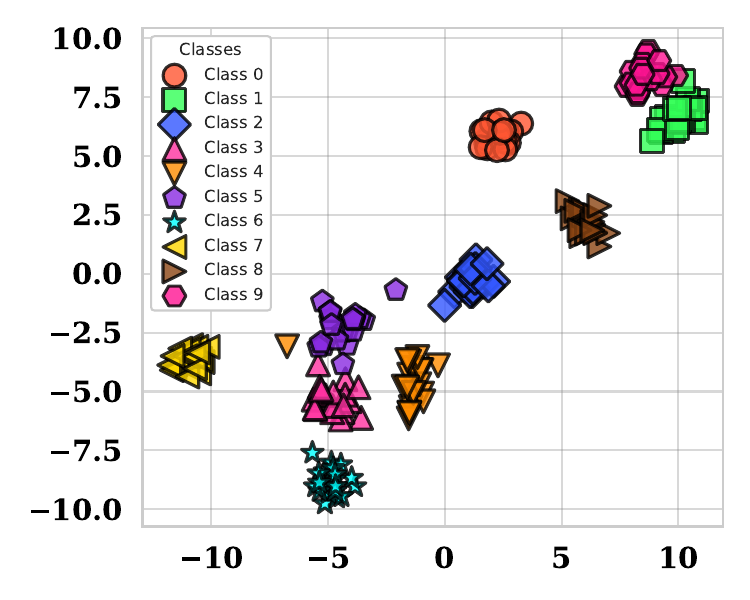}
    }
    \subfigure[FedPAC]{
    \includegraphics[width=0.23\textwidth]{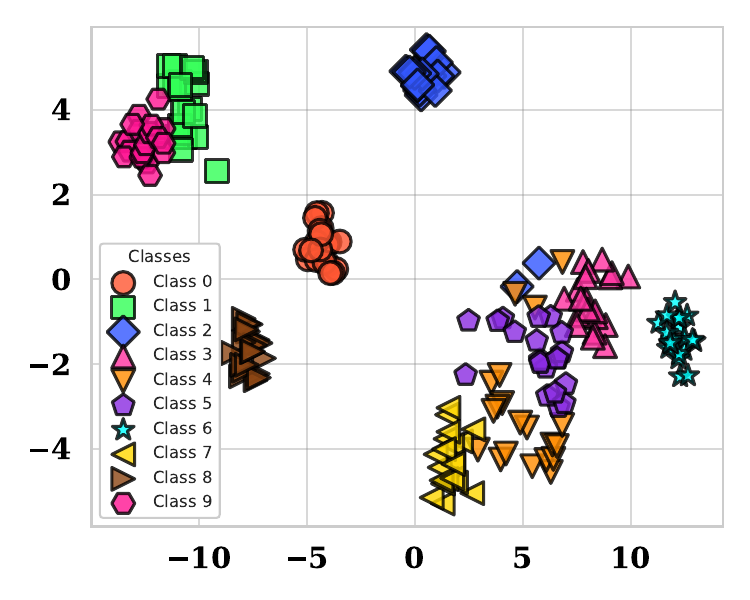}
    }
    % \hfill
    \subfigure[FedProto]{
    \includegraphics[width=0.23\textwidth]{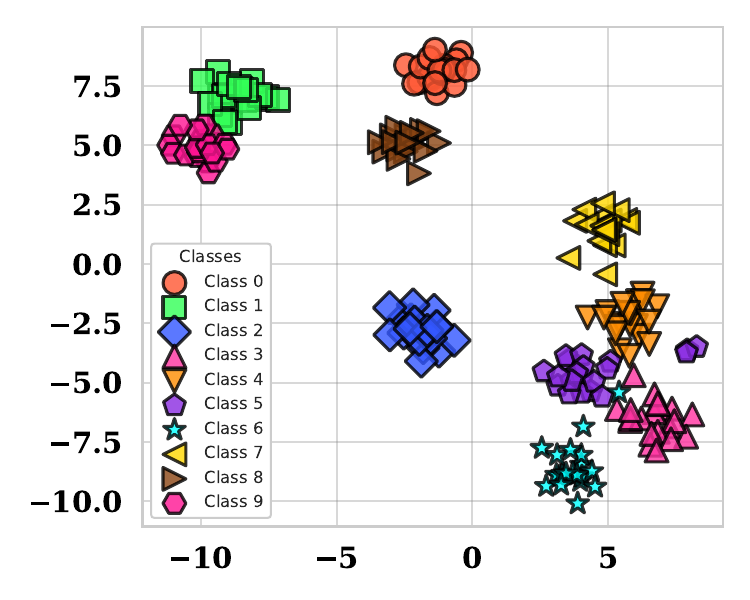}
    }
    % \hfill
    \subfigure[Ours]{
     \includegraphics[width=0.23\textwidth]{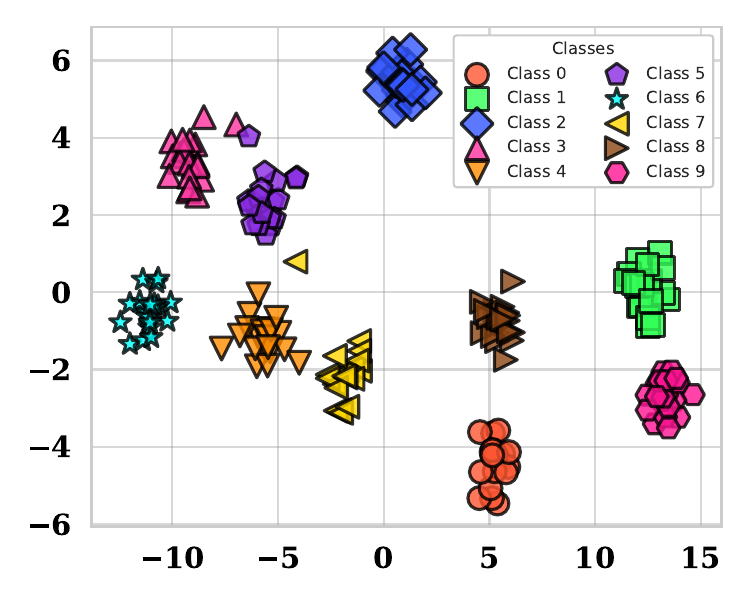}
    }
    \caption{Visualization of the local prototypes using t-SNE.}
    % \vspace{-7pt}
    \label{fig:tsne}
\end{figure}
To evaluate the stability of feature extractors and their ability to capture intrinsic class knowledge, we train models on the CINIC-10 dataset under data heterogeneity $s=20$ and visualize the local prototypes. As shown in Fig.~\ref{fig:tsne}, existing methods (FedFA, FedProto, FedPAC) exhibit prototype overlap among highly confusable classes (e.g., cat, deer, and dog), indicating suboptimal class separation in the embedding space. In contrast, our method employs a more robust global prototype construction mechanism, which enhances consistency during local training and improves the learning of intrinsic class characteristics.

\begin{figure}[!t]
    \centering
    \subfigure[Repeat ratio: 10\%]{
        \includegraphics[width=0.22\textwidth]{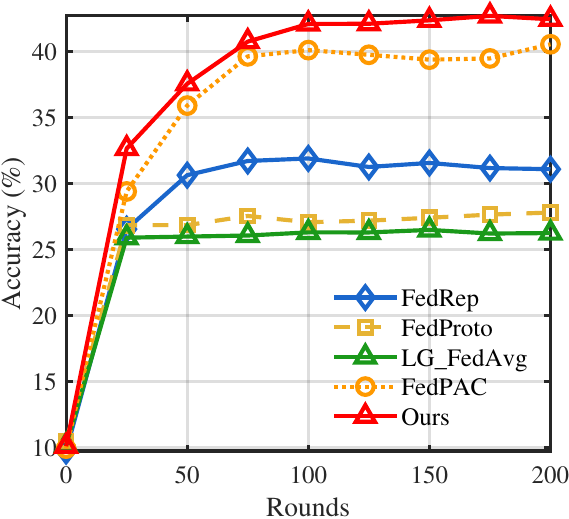}
    }
    % \hfill
    \subfigure[Repeat ratio: 25\%]{
        \includegraphics[width=0.22\textwidth]{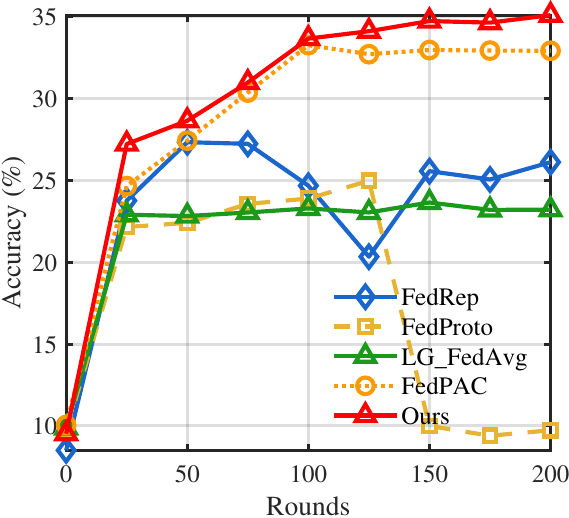}
    }
    \subfigure[Repeat ratio: 35\%]{
        \includegraphics[width=0.22\textwidth]{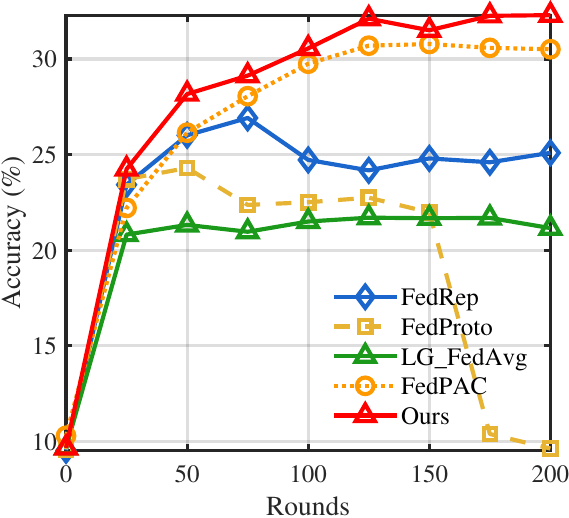}
    }
    % \hfill
    \subfigure[Repeat ratio: 50\%]{
        \includegraphics[width=0.22\textwidth]{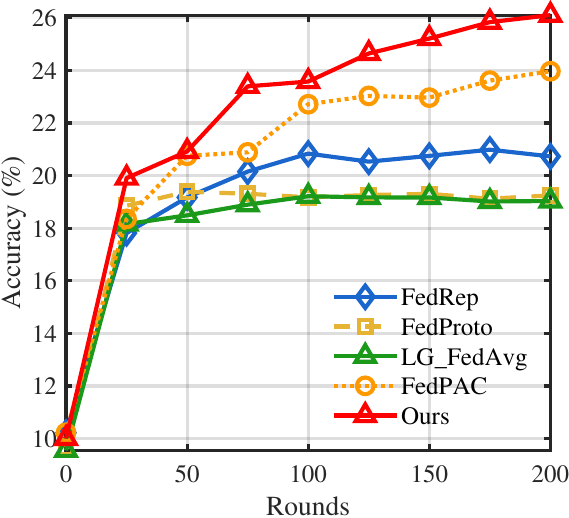}
    }
    \caption{Test accuracy with varying repeat ratios on CINIC-10 ($s=50$).}
    \label{fig:ratio}
\end{figure}

To further examine the effectiveness of the MPS component, we introduce controlled redundancy by selecting a subset of clients based on a predefined repeat ratio. The datasets of these clients contain identical samples replicated multiple times. Experimental results (Fig.~\ref{fig:ratio}) demonstrate that our method maintains stable performance across varying redundancy levels. Comparatively, FedProto and FedRep exhibit overfitting tendencies. This underscores the importance of fairly evaluating uploaded client information and forming an unbiased consensus.%, as highlighted in Section~\ref{intro}.

\textbf{Real-world scalability analysis.}
Following FedSeg, we partition the Cityscapes dataset into 152 clients, distributing 19 semantic classes such that each class is assigned to 8 clients. This setup ensures a highly heterogeneous data distribution, as each client contains data from only a single class. 

\begin{figure}[!t]
    \centering
    \includegraphics[width=0.47\textwidth]{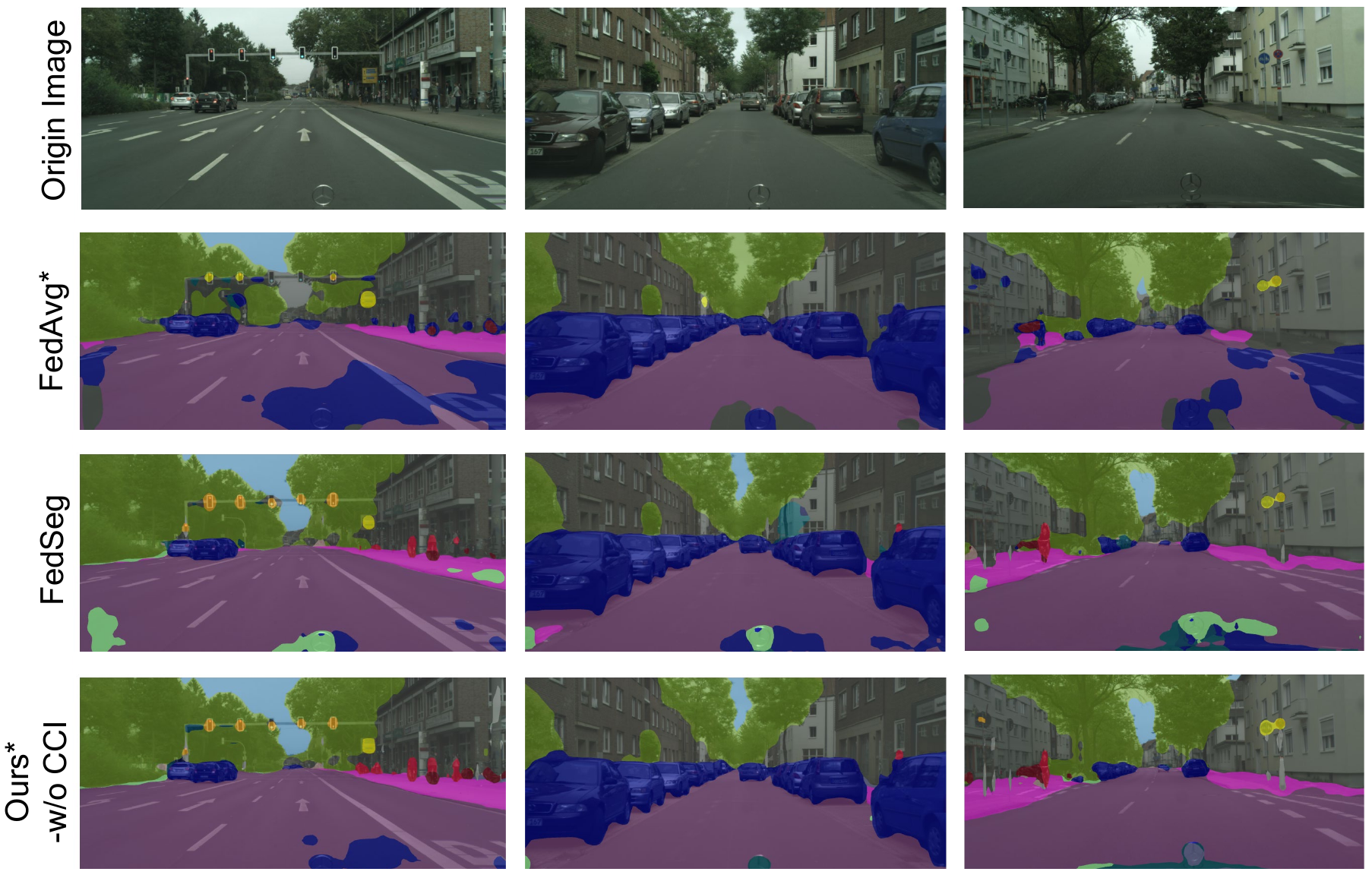}
    \caption{Semantic segmentation performance analysis. Asterisk (*) denotes methods using FedSeg-optimized cross-entropy loss; "-w/o" indicates component removed.}
    \label{fig:seg}
    % \vspace{-1pt}
\end{figure}

As shown in Fig.~\ref{fig:seg}, FedAvg struggles with certain categories (e.g.,  "person"), likely due to inadequate representation-level alignment. While FedSeg improves consistency through feature contrastive learning, it still underperforms on fine-grained or structurally complex targets. In contrast, our approach introduces a quality-aware aggregation strategy that adaptively adjusts client contributions, fostering more structured global prototypes. This enhancement yields comparatively clearer segmentation boundaries, particularly for challenging classes like "person", "building", and "traffic sign". We further evaluate our method through segmentation experiments under varying client-per-class configurations. As shown in Tab.~\ref{segclient}, our approach consistently outperforms baselines, achieving higher mean IoU (prediction-ground truth overlap) and mean accuracy scores.
% We conduct semantic segmentation training on the Cityscapes dataset with varying client numbers, where more clients per class improve performance. As shown in Tab.~\ref{segclient}, our method outperforms others in both mIoU (intersection-over-union between predicted and ground truth pixels, averaged over all classes) and mean accuracy.

% \textcolor{blue}{Additional results on convergence and further experiments are provided in Appx.~\ref{appb}.}

\begin{table}[!t]
\centering
\begin{tabular}{l|cc|cc|cc}
\toprule
\multirow{2}{*}{\textbf{Method}} & \multicolumn{2}{c|}{\textbf{Client 95}} & \multicolumn{2}{c|}{\textbf{Client 152}} & \multicolumn{2}{c}{\textbf{Client 285}} \\
                & mIoU & Acc & mIoU & Acc & mIoU & Acc \\
\midrule
FedAvg*          & 27.23 & 62.77 & 35.28 & 64.17 & 36.87  & 65.39 \\
FedSeg         & 31.27 & 66.95 & 37.33 & 68.01 & 39.19 & 68.98 \\
\textbf{Ours*}     & \textbf{32.17} & \textbf{68.79} & \textbf{39.17} & \textbf{69.91} & \textbf{40.28} & \textbf{70.69} \\
\bottomrule
\end{tabular}
\caption{Segmentation performance comparison (\%) on Cityscapes.}
\label{segclient}
\vspace{-7pt}
\end{table}

\vspace{-7pt}
\section{Conclusion}
In this paper, we propose FedMate, a novel FL framework that facilitates complementary knowledge fusion through recalibrated prototype guidance and merit-based discrimination training. Our approach achieves bilateral optimization across server and client objectives: On the server side, the MPS strategy strengthens global guidance via aggregation recalibration, while CCI refines supervisory signals to ensure consistency; on the client side, a dual-adversarial mechanism enhances both inter-class separability and cross-client generalization. Empirical results demonstrate that FedMate effectively synergizes its decoupled components, striking a superior balance between global generalization and local personalization compared to existing methods. For future work, we plan to scale FedMate to larger-scale, real-world heterogeneous scenarios, explore adaptive coordination mechanisms for decoupled components, and reduce peak communication costs per round to improve practical usability.
%for dynamic recalibration of prototype aggregation weights, enabling fine-grained classifier aggregation at the class level. Additionally, the transmission of feature extractors is conditioned on the representational capacity of prototypes, improving communication efficiency. During classifier training, a dual-adversarial mechanism is introduced to enhance both inter-class separability and cross-client generalization. Empirical results demonstrate that our method facilitates effective synergy between decoupled components, achieving a more favorable balance between global generalization and local personalization.

%In future work, we hope to scale our approach to larger-scale, real-world heterogeneous scenarios, explore adaptive and theoretically grounded mechanisms for coordinating decoupled components, and reduce peak communication costs per round to enhance usability. 

\section*{Acknowledgments}
This work was supported in part by the Taishan Scholars Program under Grants tsqn202211203 and tsqn202408239, in part by the NSFC under Grant 62402256, in part by the Shandong Provincial Nature Science Foundation of China under Grant ZR2024MF100, and in part by the QLU/SDAS Pilot Project for Integrated Innovation of Science, Education, and Industry under Grant 2024ZDZX08.

\bibliography{mybibfile}

\end{document}

% \begin{nolinenumbers}
% \thispagestyle{firstpage}

\begin{frontmatter}

\paperid{6968 ECAI-2025}  % Optional submission ID
% \title{Appendix to Choice Outweighs Effort: Facilitating Complementary Knowledge Fusion in Federated Learning via Re-calibration and Merit-discrimination}
\title{Appendix}

\end{frontmatter}
% \end{nolinenumbers}

%%%%%%%%%%%%%%%%%%%%%%%%%%%%%%%%%%%%%%%%%%%%%%%%%%%%%%%%%%%%%%%%%%%%%%%%
\vspace{-0.8cm}
\section{Algorithm Details}\label{appa}
This section provides the pseudocode for FedMate, supplementing the main text. Algorithm~\ref{alg1} describes the server-side aggregation process, while Algorithm~\ref{alg2} specifies the client-level training procedure.
% In this section, we present the pseudocode for FedMate, which supplements the main text. Algorithm~\ref{alg1} describes the server-side aggregation process, while Algorithm~\ref{alg2} specifies the client-level training procedure.
\begin{algorithm}[H]
\caption{FedMate: Server-side Aggregation Procedure}
\label{alg1}
\begin{algorithmic}[1]
\STATE \textbf{Input:} Total rounds $T$, number of clients $N$, learning rate $\eta$
\STATE Initialize global feature extractor $\theta^0$ and global classifier $\phi^0$
\FOR{each round $t = 0, 1, \dots, T-1$}
    \STATE Server selects a subset $\mathcal{Z}_t$ of clients
    \FOR{each client $i \in \mathcal{Z}_t$ \textbf{in parallel}}
        \STATE \# CT is Client-Training($\cdot$)
        \IF{$t=0$}
            \STATE $(\phi_i^{t+1}, P_i^{t+1}, \theta_i^{t+1}(\text{by } Q)) \leftarrow \text{CT}(\theta^t, \phi^t)$
        \ELSE
            \STATE $(\phi_i^{t+1}, P_i^{t+1}, \theta_i^{t+1}(\text{\text{by} }Q)) \leftarrow \text{CT}(\theta^t (\text{if have}), \phi^t, P^t)$
        \ENDIF
    \ENDFOR
    \STATE Server aggregates $\{P_i^{t+1}\}$ by Eq.~(6) to obtain $P^{t+1}$
    \STATE Server aggregates $\{\phi_i^{t+1}\}$ via Eq.~(7) and fine-tunes by Eq.~(8) to obtain $\phi^{t+1}$
    \IF{feature extractors $\{\theta_i^{t+1}\}$ are uploaded}
        \STATE Server aggregates $\{\theta_i^{t+1}\}$ weighted by $\alpha_i$ to obtain $\theta^{t+1}$
    \ENDIF
\ENDFOR
\end{algorithmic}
\end{algorithm}
\vspace{-0.7cm}
\begin{algorithm}[H]
\caption{FedMate: Client-level Training Procedure}
\label{alg2}
\begin{algorithmic}[1]
\STATE \textbf{Input:} Local epochs $E$, feature extractor $\theta^t$, classifier $\phi^t$, global prototype $P^t$, hyperparameter $\lambda_e,\lambda_c$, learning rate $\eta_l$
\IF{received $\theta^t$}
    \STATE Update local feature extractor: $\theta_i^t \leftarrow \theta^t$
\ENDIF
\STATE Compute local prototype $P_i^t$
\FOR{each local epoch $e = 1, \dots, E$}
    \STATE \# Freeze feature extractor, train classifier
    \STATE Train $\phi_i^t$ by minimizing the loss in Eq.~(3)
    \STATE \# Freeze classifier, train feature extractor
    \IF{received $P^t$}
        \STATE $\ell_{i,\text{cent}}^t=\sum_{k\in[\mathcal{C}_i]}\mathbb{L}_2(\sum_{x_{i,k}\in\mathcal{D}_{i,k}}f_i(\theta_i^t; x_{i,k}); P^t_k)$
        \STATE Train $\theta_i^t$ by minimizing $\sum_{(x_i,y_i) \in \mathcal{D}_i} \ell_{\text{ce}}(\omega_i^t; x_i,y_i) + \lambda_e\ell_{i,\text{cent}}^t$
    \ENDIF
\ENDFOR
\STATE Recompute local prototype $P_i^{t+1}$
\STATE \textbf{Return} $(\phi_i^{t+1}, P_i^{t+1}, \theta_i^{t+1}(\text{uploaded only if} \; t+1 \in Q )$
\end{algorithmic}
\end{algorithm}

\section{Additional Experimental Results}\label{appb}
\begin{figure}[H]
    \centering
    \subfigure[CIFAR-10, $s=20$]{
        \includegraphics[width=0.2\textwidth]{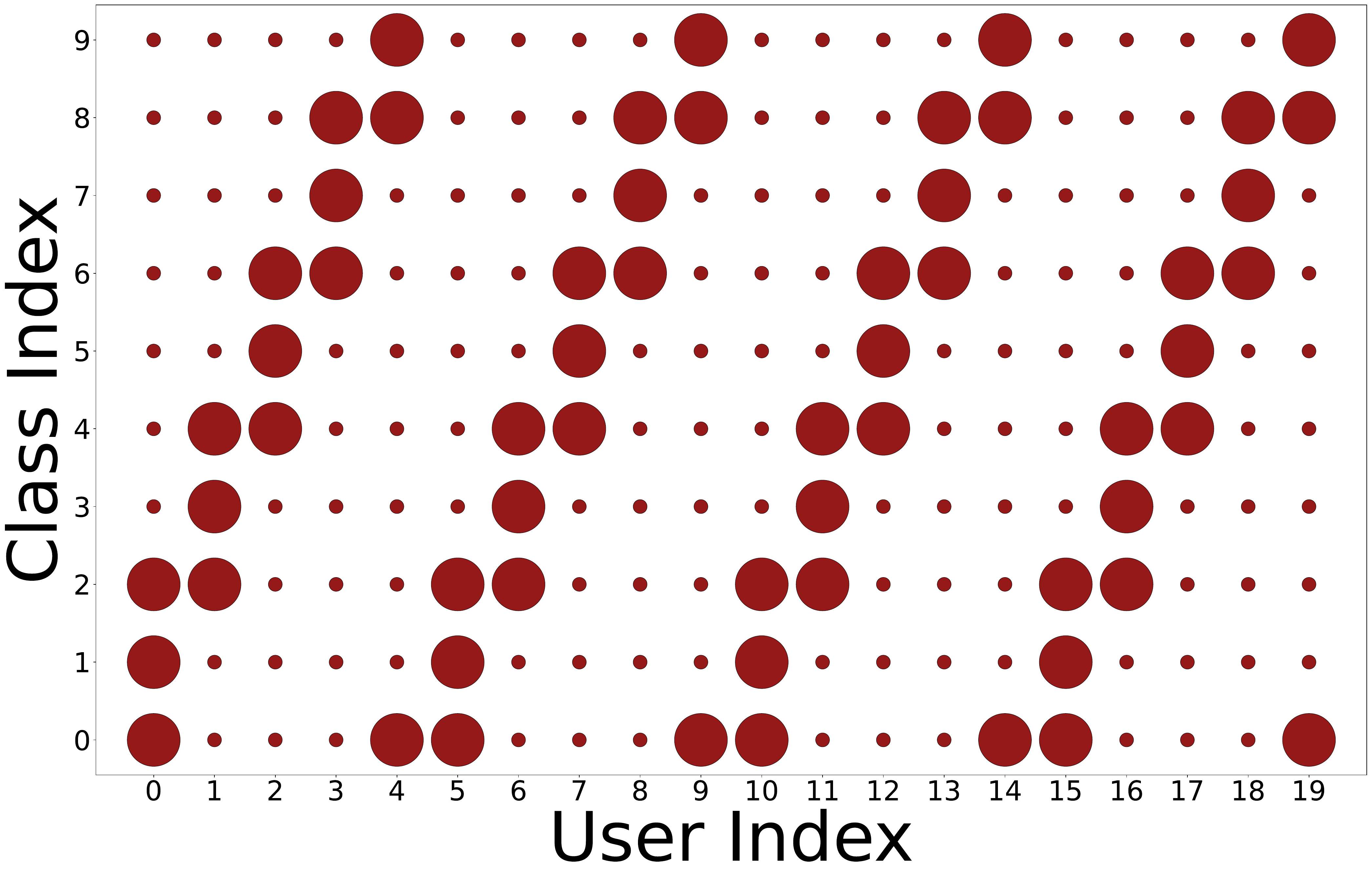}
    }
    % \hfill
    \subfigure[CIFAR-10, $s=70$]{
        \includegraphics[width=0.2\textwidth]{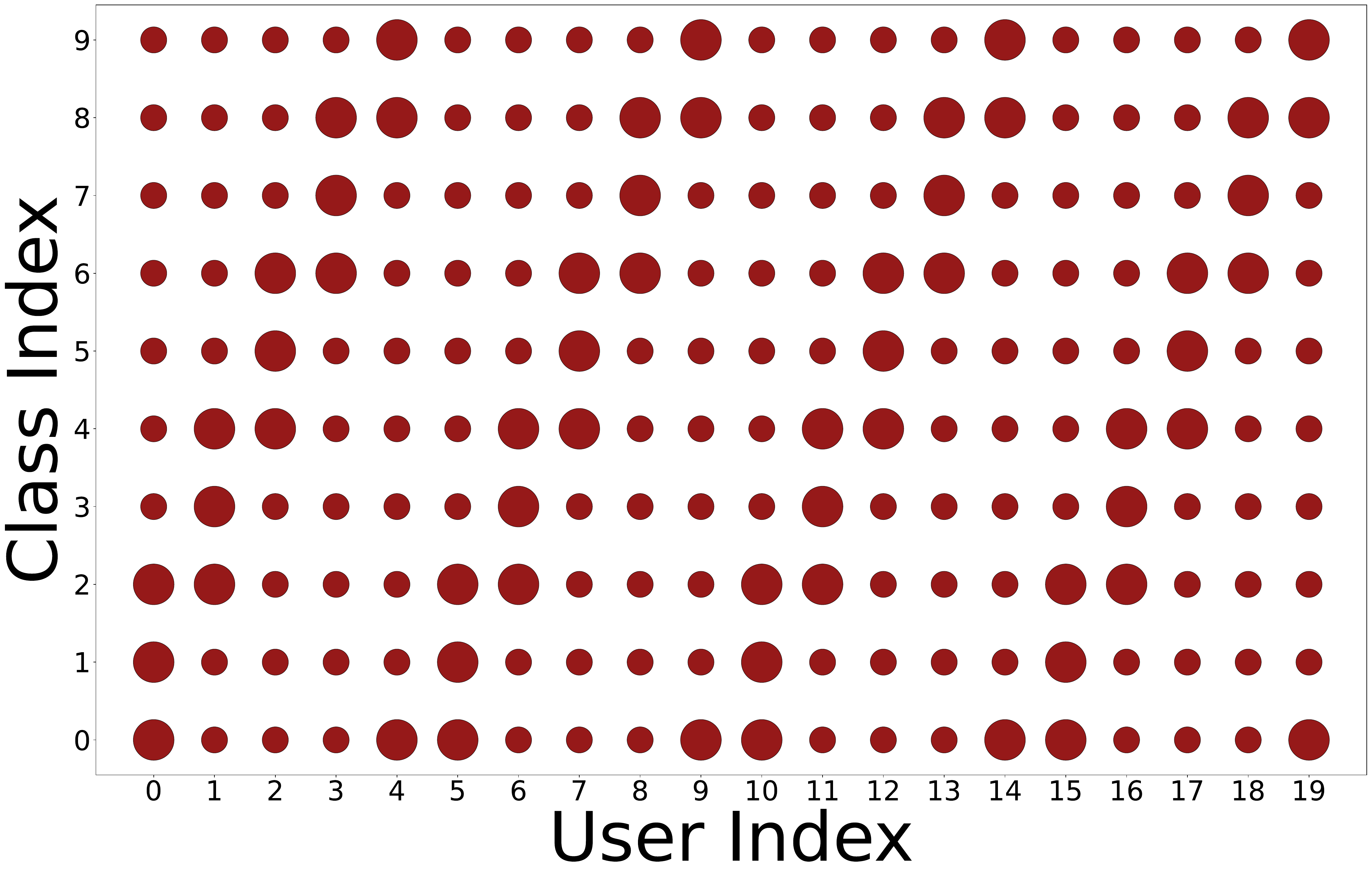}
    }
    \subfigure[CIFAR-10, Patho. Setting]{
        \includegraphics[width=0.2\textwidth]{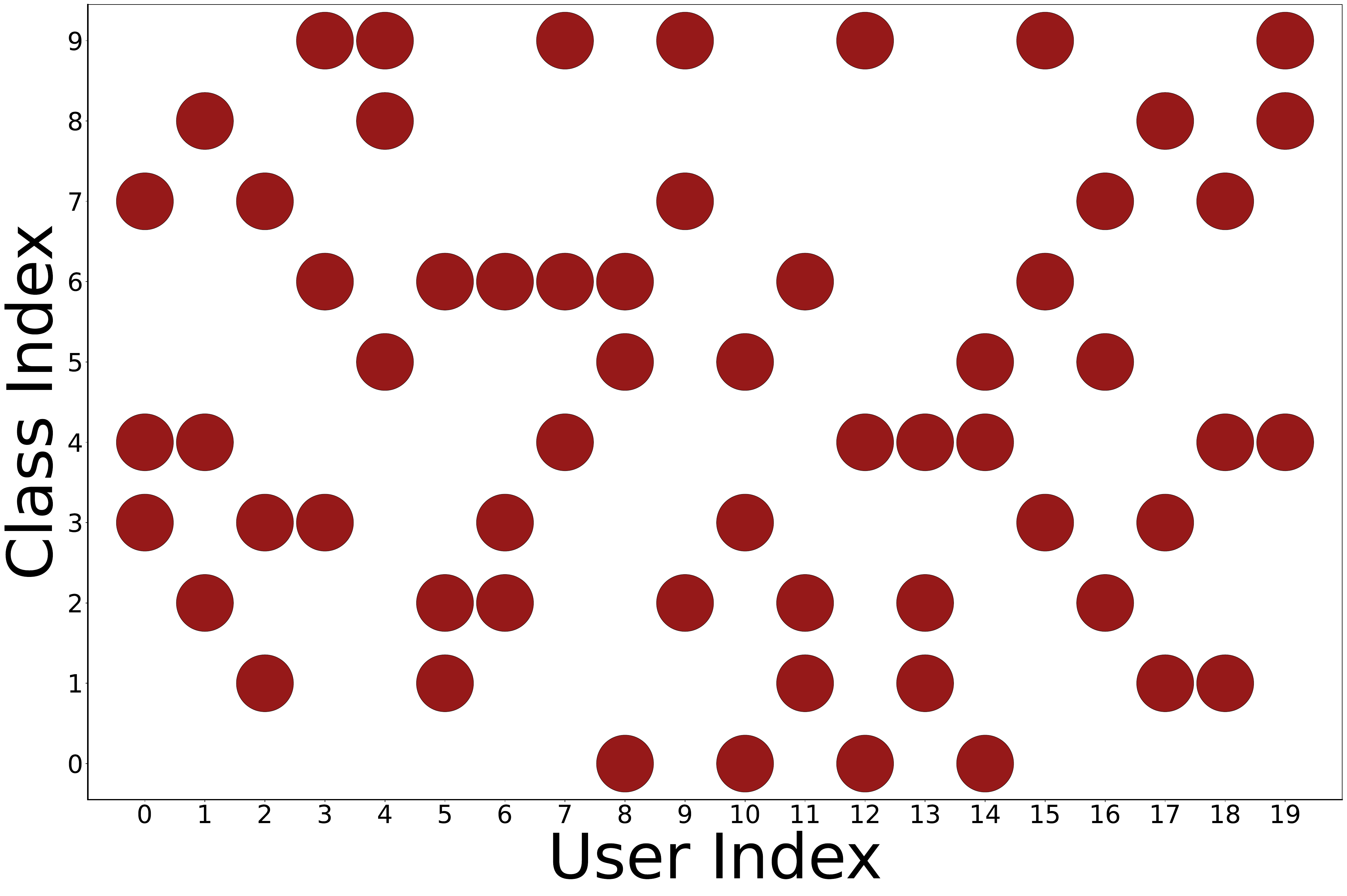}
    }
    % \hfill
    \subfigure[CIFAR-100, Patho. Setting]{
        \includegraphics[width=0.2\textwidth]{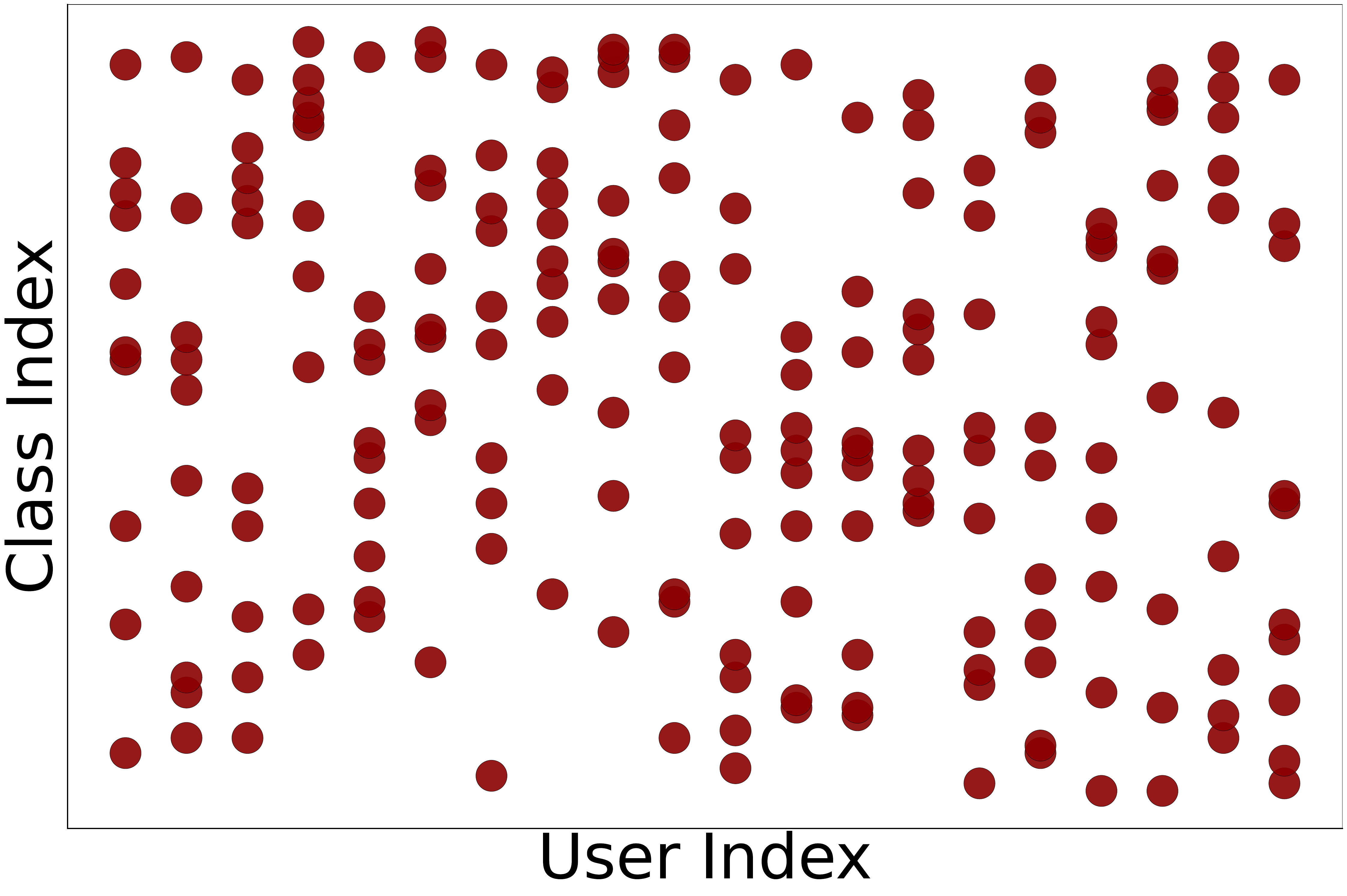}
    }
    \vspace{-10pt}
    \caption{Data distribution under different heterogeneous scenarios.}
    \label{Data}
    \vspace{-0.5cm}
\end{figure}
\textbf{Data distribution visualization.} To better characterize the heterogeneous learning scenarios under investigation, Fig.~\ref{Data} illustrates the data distributions in extreme pathological (Patho.) settings across different datasets, and under varying degrees of skewness ($s\%$).
% with varying degrees of skewness ($s\%$).

\begin{figure}[H]
    \centering
    \subfigure[$s=70$]{
        \label{cft1}
        \includegraphics[width=0.2\textwidth]{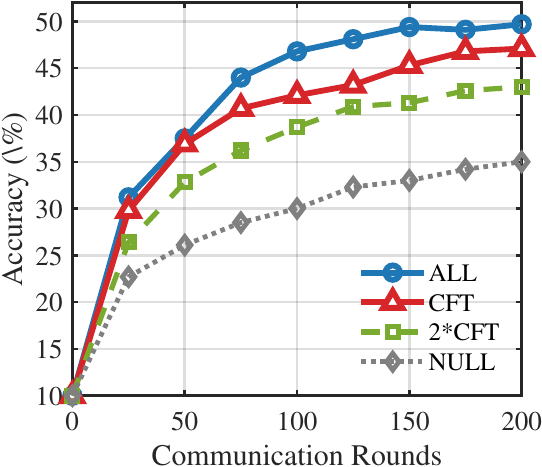}
    }
    \subfigure[Patho. Setting]{
        \label{cft2}
        \includegraphics[width=0.205\textwidth]{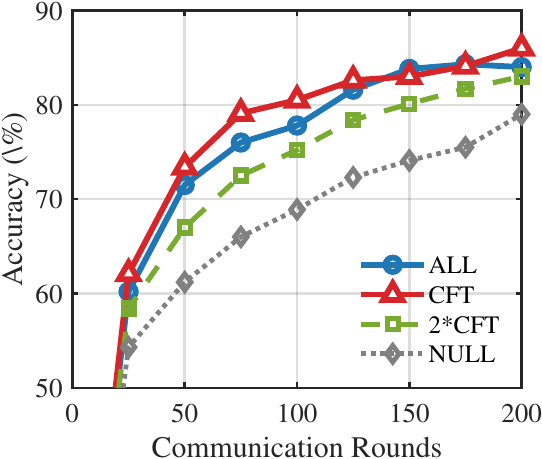}
    }
    \vspace{-10pt}
    \caption{Accuracy curves under different feature extractor transmission strategies: {ALL} transmits every round, {CFT} follows the predefined $Q$, {2*CFT} transmits half of $Q$, and {NULL} disables transmission.}
    \vspace{-0.5cm}
\end{figure}

\textbf{CFT component evaluation.} We evaluate our approach on the CINIC-10 dataset under two conditions: 1) a uniform test distribution (training: $s=70$) and 2) a heterogeneous test distribution matching the training set (training: Patho.). As shown in Fig.~\ref{cft1} and Fig.~\ref{cft2}, CFT effectively balances model performance with communication efficiency by strategically uploading local feature extractors at appropriate rounds. 
%We conduct training on the CINIC-10 dataset with $s=70$ / Patho. setting and evaluate our models on two types of test sets: one uniformly distributed across all classes and another matching the non-IID distribution of the training set. The experimental results are shown in Fig.~\ref{cft1} and Fig.~\ref{cft2}. As observed, CFT achieves a balance between model effectiveness and communication overhead by uploading local feature extractors at appropriate rounds. 
The results demonstrate that CFT achieves superior performance on the heterogeneous test set (which better reflects real-world local data distributions) compared to transmitting feature extractors in every round (ALL). This improvement stems from CFT's ability to relax strict global generalization requirements, thereby achieving a more favorable balance between generalization and personalization. Notably, the selective transmission mechanism based on the predefined threshold $Q$ reduces communication overhead while maintaining model effectiveness.
%Notably, when evaluated on the non-IID test set aligned with the training distribution, which better reflects local adaptability, selectively transmitting the feature extractor using the predefined $Q$ (CFT) outperforms the approach of transmitting the feature extractor in every round (ALL). This relaxation of strong global generalization constraints enables the model to better balance the trade-off between generalization and personalization.

\begin{figure}[H]
    \centering
    \subfigure[CIFAR-10]{
        \includegraphics[width=0.23\textwidth]{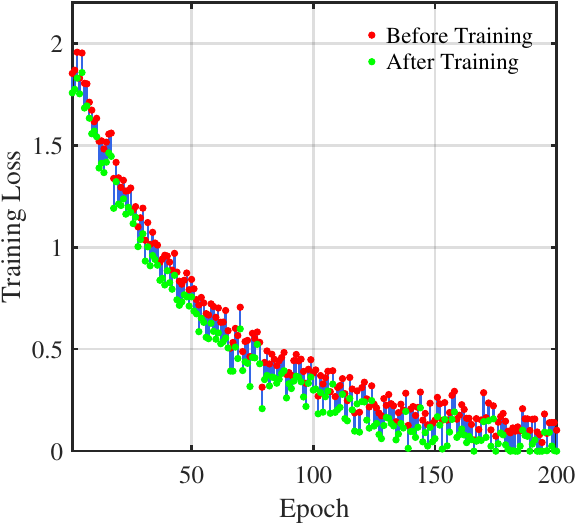}
    }
    \subfigure[CIFAR-100]{
        \includegraphics[width=0.23\textwidth]{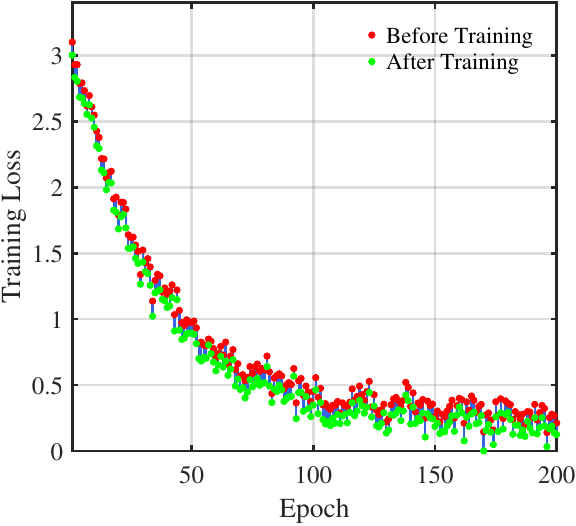}
    }
    \subfigure[CINIC-10]{
        \includegraphics[width=0.23\textwidth]{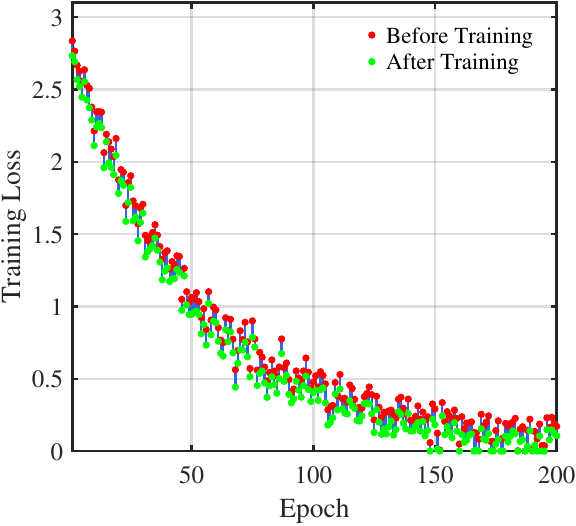}
    }
    \subfigure[EMNIST]{
        \includegraphics[width=0.23\textwidth]{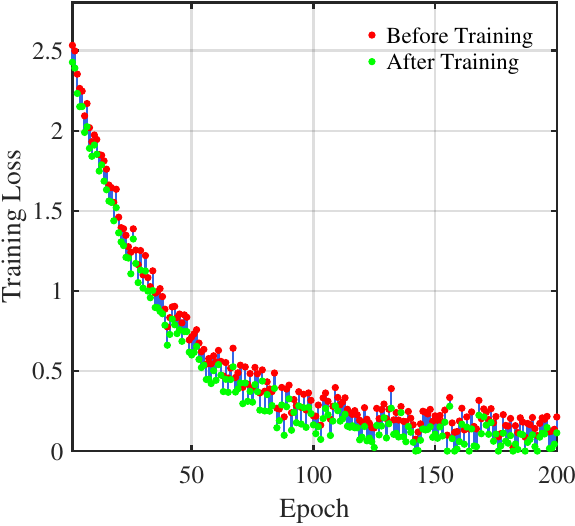}
    }
    \vspace{-10pt}
    \caption{Training loss curves under different dataset.}
    \label{loss}
\end{figure}

\textbf{Convergence analysis.} 
We examine the convergence properties of our method under heterogeneous training conditions ($s=20$) across datasets with varying complexity levels. Fig.~\ref{loss} presents the training dynamics, comparing model loss before and after local updates. The results demonstrate that our approach achieves consistent and stable convergence within approximately 150 communication rounds for all evaluated datasets, indicating robust performance regardless of dataset difficulty. 
%We analyze the convergence behavior of our method across datasets of varying difficulty, configured with a heterogeneous training setup at $s=20$. We measure the model’s loss before and after local training, with the results summarized in Fig.~\ref{loss}. As shown, our method consistently achieves stable convergence in approximately 150 rounds across all datasets.
\vspace{-10pt}

\textbf{System communication cost.} 
We conduct a comprehensive evaluation of communication overhead across baseline methods using identical training rounds. As presented in Tab.~\ref{cost}, while lightweight approaches (FedProto, LG-FedAvg, FedPer, and FedRep) incur lower transmission costs compared to conventional FL, their performance remains constrained to specific heterogeneous scenarios. Notably, our method maintains the communication footprint of model-only transmission while achieving competitive accuracy comparable to information-rich alternatives. 
%We evaluate the total communication overhead of baselines under the same number of training rounds. As shown in Tab.~\ref{cost}, methods like FedProto, LG-FedAvg, FedPer, and FedRep incur lower transmission costs than traditional FL, but their effectiveness is often limited to specific heterogeneous scenarios. Our method matches the communication costs of model-only transmission, while delivering comparable performance to methods transmitting more information.
\begin{table}[H]
\centering
\caption{System communication cost comparison.}
\label{cost}
\begin{tabular}{lc}
\toprule
\textbf{Method} & \textbf{Transmission} \\ 
\midrule 
FedProto       &Prototype  \\
LG-FedAvg      &Classifier  \\ 
FedPer         &Feature Extractor  \\ 
FedRep         &Feature Extractor  \\ 
FedBABU        &Feature Extractor  \\ 
FT-FedAvg      &Model  \\ 
MOON           &Model  \\ 
FedFA          &Model+Prototype  \\ 
FedPAC         &Model+Prototype  \\ 
\textbf{Ours}           &Equivalent to \textbf{Model}  \\
\bottomrule
\end{tabular}
\end{table}

\textbf{Hyperparameter analysis.}
Our framework introduces two key hyperparameters: 1) $\lambda_e$ for local feature extractor training and 2) $\lambda_c$ for local classifier adaptation. We systematically evaluate our approach on the CINIC-10 dataset ($s = 70$) with uniform test distribution. Tabs.~\ref{lambda_e} and~\ref{lambda_c} demonstrate that extreme values in either direction--excessive or insufficient reliance on global knowledge--respectively cause overfitting or underfitting. The optimal balance emerges at $\lambda_e = 0.8$ and $\lambda_c = 0.6$. 

%Our method introduces only two hyperparameters: $\lambda_e$ and $\lambda_c$, which respectively govern the training of the local feature extractor and local classifier. We conduct experiments on the CINIC-10 dataset with $s = 70$ and evaluate the model on uniformly distributed test data. As shown in Tab.~\ref{lambda_e} and Tab.~\ref{lambda_c}, the results indicate that both overusing and underusing global information can lead to model underfitting or overfitting. The optimal performance is achieved when $\lambda_e = 0.8$ and $\lambda_c = 0.6$.
% \vspace{-10pt}
\begin{table}[H]
\centering
\small
\begin{tabular}{c|cccc}
\toprule
\textbf{$\lambda_e$} & 0.0 & 0.2 & 0.5 & 0.8 \\
\midrule
\textbf{Accuracy} & 47.57 & 48.62 & 49.85 & \textbf{50.27} \\
\midrule
\textbf{$\lambda_e$} & 1.0 & 2.0 & 5.0 & 7.0 \\
\midrule
\textbf{Accuracy} & 49.94 & 48.12 & 43.56 & 40.29 \\
\bottomrule
\end{tabular}
\caption{Test accuracy under different values of $\lambda_e$.}
\label{lambda_e}
\end{table}
\vspace{-18pt}
\begin{table}[H]
\centering
\small
\begin{tabular}{c|cccc}
\toprule
\textbf{$\lambda_c$} & 0.0 & 0.2 & 0.4 & 0.6 \\
\midrule
\textbf{Accuracy} & 45.68 & 46.77 & 48.28 & \textbf{49.97} \\
\midrule
\textbf{$\lambda_c$} & 0.8 & 1.0 & 2.0 & 3.0 \\
\midrule
\textbf{Accuracy} & 49.21 & 48.98 & 46.38 & 44.50 \\
\bottomrule
\end{tabular}
\caption{Test accuracy under different values of $\lambda_c$.}
\label{lambda_c}
\end{table}

\vspace{-13pt}
\textbf{Client computational efficiency analysis.}
We measure overhead by the wall-clock time to reach 70\% accuracy on CINIC-10 (Pathology). Our method requires only 17.7 minutes, compared to 18.7–25.3 minutes for baselines; relative to the median baseline (20.85 minutes), this represents a 3.15-minute (15\%) speedup. While the feature extractor and classifier are decoupled during local training, consistency between global prototypes and class-level knowledge aligns their objectives, enabling faster convergence with balanced generalization and personalization. The additional discriminator is a lightweight binary classifier, incurring only modest overhead.
\begin{figure}[H]
    \centering
    \includegraphics[width=0.32\textwidth]{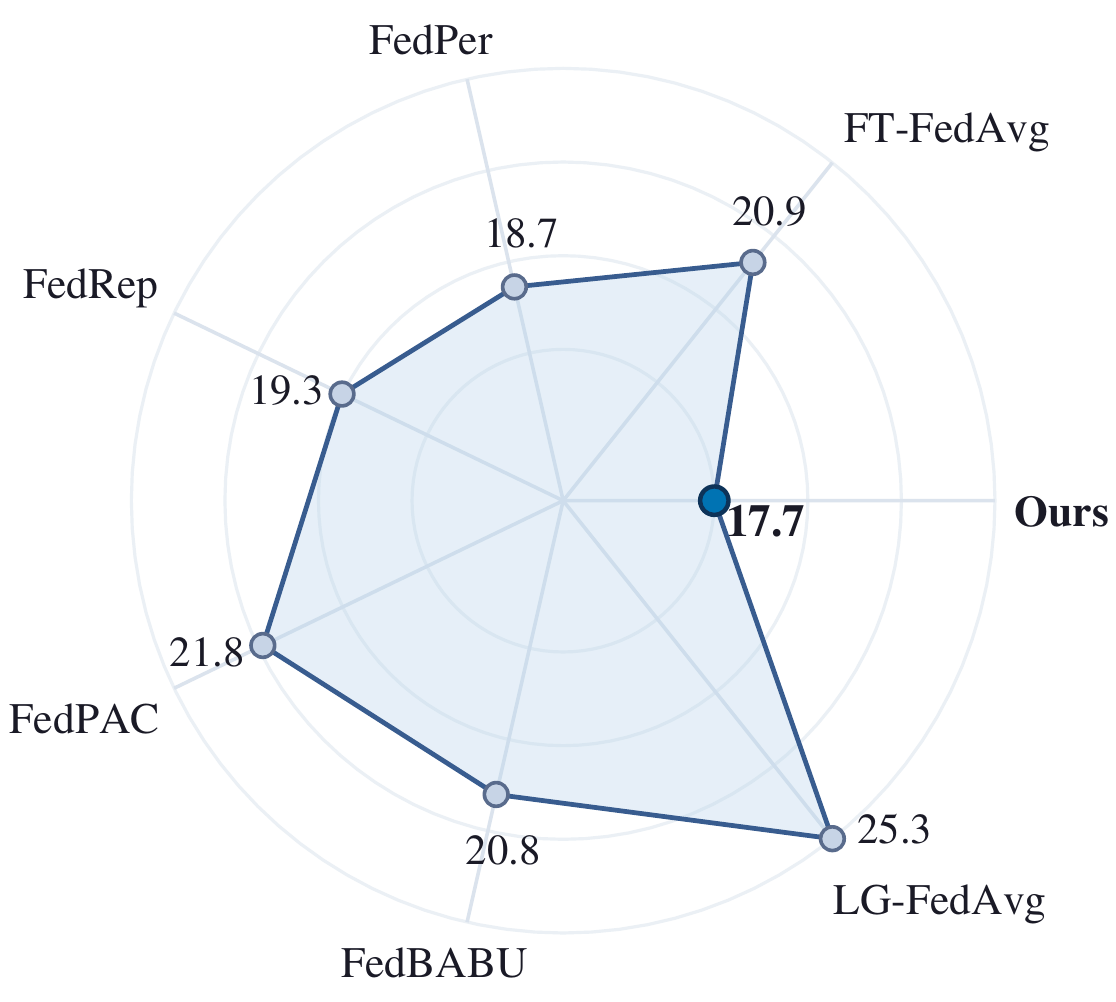}
    \caption{Client computation overhead (minutes).}
    \label{overhead}
\end{figure}
\vspace{-0.6cm}
\textbf{Single component ablation analysis.}
To assess the contribution of each component, we conduct ablation studies by isolating them individually (Tab.~\ref{single}). The results demonstrate that all components provide consistent performance improvements, further underscoring their effectiveness. Moreover, by cross-referencing with Tab.~1 in the main text, we observe that components related to feature extraction and classification decision exhibit strong synergy when combined. In addition, when both feature-level and classifier-level optimization are applied simultaneously, they complement each other and further enhance the consistency optimization process.
\begin{table}[ht]
\label{single}
\centering
\caption{Single component ablation ($s=70$).}
\begin{tabular}{l@{\hskip 12pt}c@{\hskip 12pt}c@{\hskip 12pt}c}
\toprule
\textbf{Method} & \textbf{CINIC-10} & \textbf{CIFAR-10} & \textbf{EMNIST} \\
\midrule
Backbone & $34.73 \pm 0.2$ & $54.20 \pm 0.1$ & $55.37 \pm 0.1$ \\
CFT      & $42.13 \pm 0.2$ & $59.67 \pm 0.2$ & $63.00 \pm 0.2$ \\
MPS      & $40.19 \pm 0.2$ & $64.37 \pm 0.1$ & $63.57 \pm 0.1$ \\
CCI      & $37.55 \pm 0.1$ & $61.33 \pm 0.1$ & $62.67 \pm 0.1$ \\
CCF      & $41.76 \pm 0.2$ & $63.89 \pm 0.1$ & $64.53 \pm 0.1$ \\
\textbf{FedMate}  & \textbf{$50.24 \pm 0.1$} & \textbf{$72.10 \pm 0.1$} & \textbf{$71.98 \pm 0.1$} \\
\bottomrule
\end{tabular}
\end{table}

\vspace{-15pt}

\begin{figure}[H]
\label{fig3}
    \centering
    \includegraphics[width=0.45\textwidth]{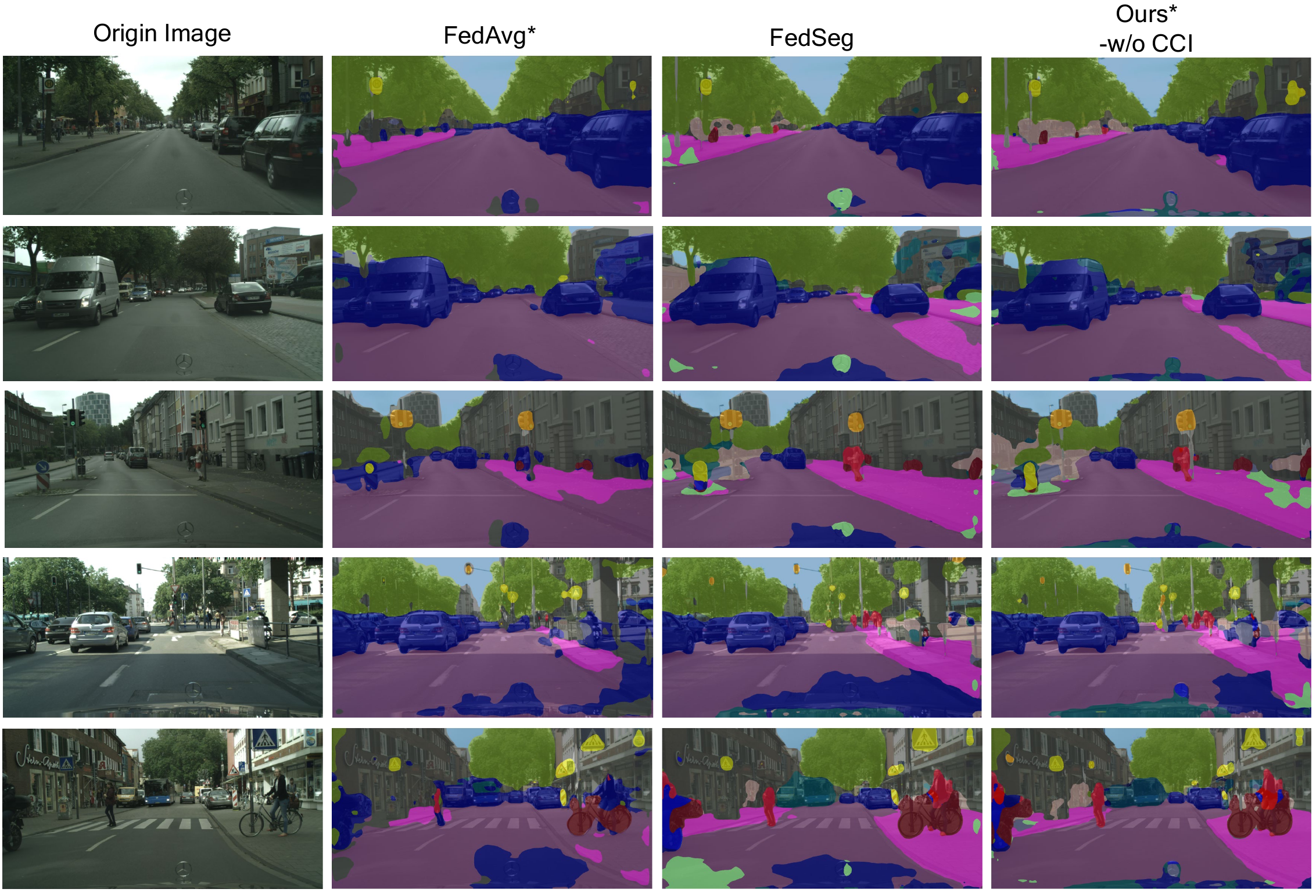}
    \vspace{-5pt}
    \caption{Semantic segmentation results on Cityscapes.}
    \label{aseg}
\end{figure}

\textbf{Additional semantic segmentation results.}
We extend our evaluation to large-scale semantic segmentation using the Cityscapes dataset under extreme heterogeneity conditions (152 clients). Fig.~\ref{aseg} presents qualitative comparisons showing our prototype-recalibration approach consistently outperforms baseline methods in two key aspects: 1) superior visual coherence in predicted segmentation maps, and 2) more precise retention of structural boundaries. These results validate the operational efficacy of our global prototype recalibration mechanism in challenging FL environments.

\begin{figure}[h]
\vspace{-3pt}
\centering
\includegraphics[width=0.80\linewidth]{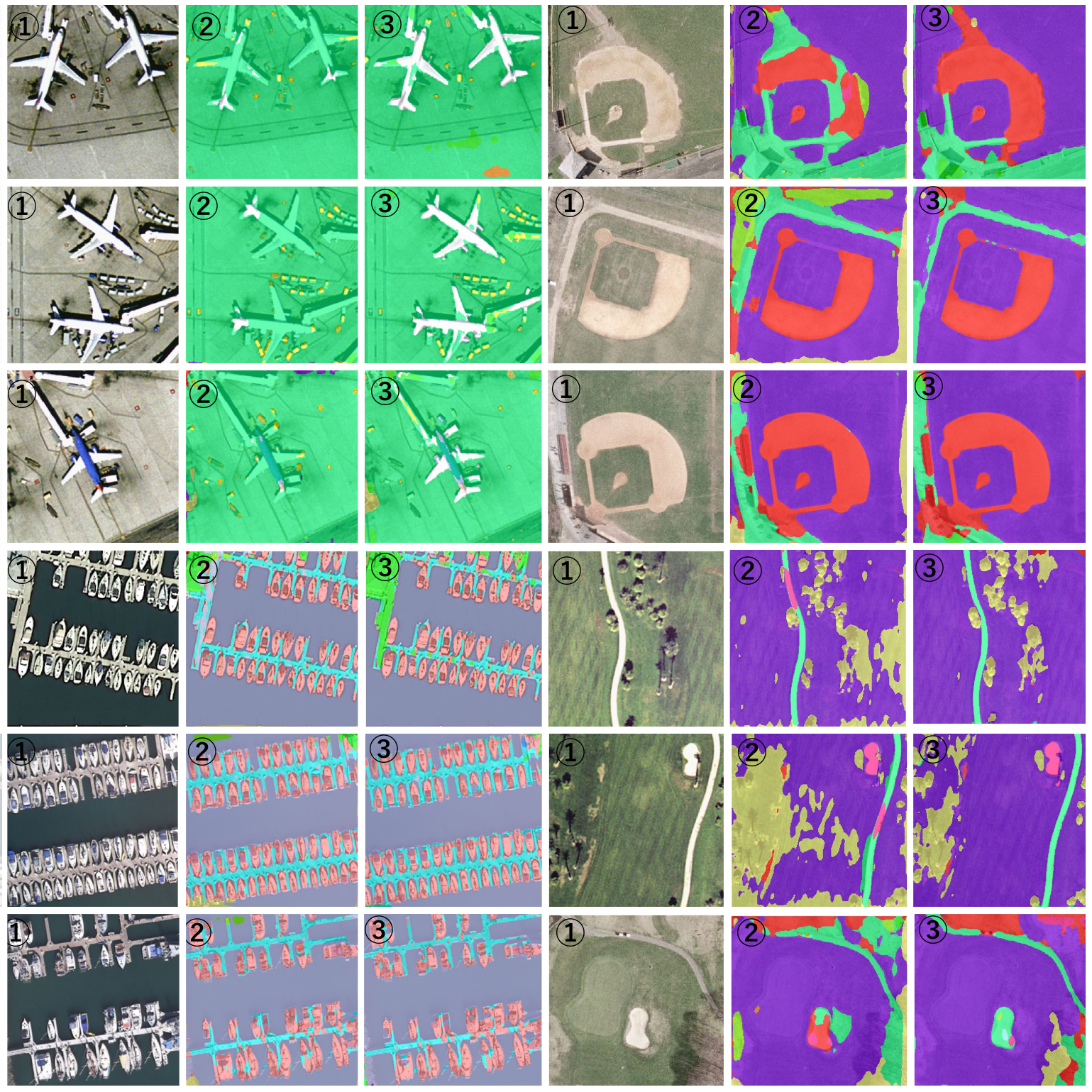}
\vspace{-5pt}
\caption{Segmentation on PatternNet: \ding{172} Original; \ding{173} FedSeg; \ding{174} FedMate.}
\label{fig:pattern}
\vspace{-5pt}
\end{figure}

In addition, we evaluate our method on the PatternNet dataset, which consists of high-resolution remote sensing imagery. The results (Fig.~\ref{fig:pattern}) show that our approach yields more fine-grained segmentation, particularly for small or structurally complex categories such as cars, ships, and roads, thereby further demonstrating its robustness across diverse domains and data characteristics.
%As illustrated in Fig.~\ref{aseg}, our method, leveraging recalibrated global prototypes, clearly surpasses baseline approaches by generating segmentation maps with higher visual fidelity and more accurate preservation of structural details. This demonstrates the practical effectiveness of our proposed recalibration mechanism under challenging federated scenarios.
% We conduct semantic segmentation training on the Cityscapes dataset with varying client numbers, where more clients per class improve performance. As shown in Tab.~\ref{segclient}, our method outperforms others in both mIoU (intersection-over-union between predicted and ground truth pixels, averaged over all classes) and mean accuracy. Fig.~\ref{aseg} further shows that our approach, using recalibrated global prototypes, surpasses the baselines in visual quality and detail preservation.

% \begin{table}[H]
% \centering
% \begin{tabular}{l|cc|cc|cc}
% \toprule
% \multirow{2}{*}{\textbf{Method}} & \multicolumn{2}{c|}{\textbf{Client 95}} & \multicolumn{2}{c|}{\textbf{Client 152}} & \multicolumn{2}{c}{\textbf{Client 285}} \\
%                 & mIoU & Acc & mIoU & Acc & mIoU & Acc \\
% \midrule
% FedAvg*          & 27.23 & 62.77 & 35.28 & 64.17 & 36.87  & 65.39 \\
% FedSeg         & 31.27 & 66.95 & 37.33 & 68.01 & 39.19 & 68.98 \\
% \textbf{Ours*}     & \textbf{32.17} & \textbf{68.79} & \textbf{39.17} & \textbf{69.91} & \textbf{40.28} & \textbf{70.69} \\
% \bottomrule
% \end{tabular}
% \caption{Accuracy with varying client numbers on Cityscapes}
% \label{segclient}
% \end{table}
% \vspace{-5pt}
\vspace{-0.5cm}

% \vspace{-0.2cm}
% \begin{quote}
% \textbf{Code repository:} \url{https://github.com/Dongrun-Li/FedMate.git}
% \end{quote}

\section{Additional Functional Details}
\textbf{Effectiveness on segmentation tasks.}
Segmentation can be regarded as a fine-grained extension of classification, where the goal is to perform pixel-level labeling. The success of both classification and segmentation relies on a high-quality feature space, in which intra-class samples are compact while inter-class samples are well separated. Constructing such a feature space requires feature extractors that balance generalization with discrimination. Our method addresses this by incorporating multi-view perspectives, leading to fairer and more robust feature aggregation and yielding an unbiased global prototype. This prototype constrains local feature extractors during training, thereby enhancing the overall feature space. As a result, our approach provides a solid foundation for downstream tasks, including classification, segmentation, and object detection.

\textbf{Workflow of the CCF module.} 
The CCF component employs two discriminators: a prototype discriminator $D_{pr,i}$ and a classification discriminator $D_{cl,i}$. The prototype discriminator takes local and global prototypes as input through the local classifier $\phi_i$ and is trained to classify local outputs as 0, while $\phi_i$ is optimized to produce outputs classified as 1, forming an adversarial process. Similarly, the classification discriminator $D_{cl,i}$ receives the global prototype through both local $\phi_i$ and global classifiers, distinguishing $\phi_i$’s outputs as 0 while forcing $\phi_i$ to generate outputs closer to 1. To mitigate forgetting, local prototypes are also used to construct a cross-entropy loss. By integrating adversarial learning with prototype supervision, the objective reduces class-specific bias, improves generalization, and alleviates catastrophic forgetting, thereby promoting stronger synergy among complementary knowledge.